\documentclass[preprint,12pt]{elsarticle}

\usepackage{graphicx}
\usepackage{amssymb}
\usepackage{booktabs}

\usepackage{tabularx}
\usepackage{lineno}
\usepackage{graphicx}
\usepackage{placeins}
\usepackage[utf8]{inputenc}
\usepackage{lmodern}
\usepackage[T1]{fontenc}
\usepackage{lscape}
\usepackage{amsmath}	
\usepackage{amssymb}
\usepackage{cases}
\usepackage{graphicx}
\usepackage{caption}
\usepackage{comment}

\usepackage{makecell}
\usepackage{multirow}
\usepackage{amsfonts}
\usepackage{float} 
\usepackage{diagbox}
\usepackage{fullpage}
\usepackage{url}
\usepackage{rotating}
\usepackage{eurosym}
\usepackage{wrapfig}
\usepackage[final]{pdfpages} 
\usepackage{epstopdf} 
\usepackage[a4paper]{geometry}
\usepackage[nottoc]{tocbibind}
\usepackage{placeins}
\usepackage{float}
\usepackage{listings}
\usepackage{color, colortbl}
\usepackage{hyperref}
\usepackage{xcolor}
\usepackage{graphicx, caption, subcaption}

\usepackage{sidecap}
\hypersetup{
colorlinks,
citecolor=black,
filecolor=black,
linkcolor=black,
urlcolor=black
} 

\journal{Journal Comput. Methods Appl. Mech. Engrg.}

\begin{document}

\begin{frontmatter}


\title{Trajectory-Aware Flow Matching for Topology Optimisation}

\author[aff1]{Shusheng Xiao}

\author[aff2]{Jinshuai Bai\corref{cor1}}
\ead{bjs@mail.tsinghua.edu.cn}

\author[aff1]{Hyogu Jeong}

\author[aff4,aff1]{Yunfei Xi}

\author[aff3]{Yilin Gui}

\author[aff1]{YuanTong Gu}

\cortext[cor1]{Corresponding author}

\affiliation[aff1]{
    organization={School of Mechanical, Medical and Process Engineering, Queensland University of Technology},
    city={Brisbane},
    postcode={4000},
    state={QLD},
    country={Australia}
}

\affiliation[aff2]{
    organization={Institute of Biomechanics and Medical Engineering, AML, Department of Engineering Mechanics, Tsinghua University},
    city={Beijing},
    postcode={100084},
    country={China}
}

\affiliation[aff3]{
    organization={School of Civil and Environmental Engineering, Queensland University of Technology},
    city={Brisbane},
    postcode={4000},
    state={QLD},
    country={Australia}
}

\affiliation[aff4]{
    organization={State Key Laboratory of Advanced Environmental Technology, Guangzhou Institute of Geochemistry, Chinese Academy of Sciences},
    city={Guangzhou},
    postcode={510640},
    country={China}
}


\begin{abstract}
Topology optimisation (TO) often requires repeated finite element analysis and
sensitivity-based material updates, which can be costly when multiple candidate designs are
needed under varying physical and design conditions. Generative TO provides a route to rapid
design exploration, but existing generative models may require adversarial training, long
reverse-diffusion sampling, or external guidance to maintain structural feasibility and physical
consistency. This study develops a flow matching-based topology optimisation (FMTO)
framework for conditional topology generation. Linear FMTO is first formulated by
interpolating between a Gaussian source field and the BESO reference topology, establishing an
efficient endpoint-based baseline but also revealing the limitation of probability paths
constructed only from the source and final topology. To address this limitation, a
trajectory-aware FMTO formulation is proposed, in which volume-fraction-indexed intermediate
BESO states are used to construct both the probability path and the target velocity field. In
this way, physics-guided optimisation history is incorporated into the learned generative flow
without adding inference-time optimisation. 
The role of trajectory guidance is analysed through a path--velocity mismatch formulation,
which explains why moderate trajectory weighting can improve generation stability, whereas
excessive guidance may over-constrain the learned transport. Numerical examples show that
linear FMTO generates diverse topology candidates with improved compliance-related
performance, volume-fraction satisfaction, topology fidelity, and substantially fewer sampling
steps than a diffusion-based baseline. Under limited training data, trajectory-aware FMTO
achieves the best overall performance with a moderate trajectory weight, consistent with the
theoretical analysis. Further studies on trajectory-anchor density and three-dimensional
topology generation demonstrate the influence of path design and the applicability of the
proposed framework beyond two-dimensional problems.

\end{abstract}


\end{frontmatter}

\clearpage


\section{Introduction}

Topology optimisation (TO) provides a systematic method for generating material layouts that satisfy prescribed loading, boundary, and design constraints while optimising structural performance~\cite{rozvany2001aims}. Over the past decades, a wide range of TO formulations have been developed, including density-based methods such as solid isotropic material with penalisation (SIMP)~\cite{bendsoe1989optimal,zhou1991coc}, evolutionary approaches such as evolutionary structural optimisation (ESO) and bi-directional evolutionary structural optimisation (BESO)~\cite{xie1993simple,querin1998evolutionary}, level-set-based methods~\cite{wang2003level}, and explicit geometry-based methods such as moving morphable components (MMC)~\cite{zhang2016new}. These methods have enabled the design of lightweight and mechanically efficient structures in aerospace, mechanical, and civil engineering applications~\cite{sigmund2013topology,rozvany2009critical,zhu2016topology}. However, conventional TO is usually performed as an iterative and problem-specific optimisation process, where repeated finite element analyses (FEA) and sensitivity evaluations are required for each new loading, boundary-condition, or constraint setting. This repeated optimisation can become computationally expensive when multiple design alternatives must be generated under changing design requirements~\cite{behzadi2021real}. These limitations have motivated increasing interest in machine learning-based TO (MLTO) methods for accelerating topology generation from prescribed design conditions.

Recent advances in ML have enabled surrogate-based topology prediction, where neural networks are trained to approximate the mapping from prescribed design conditions, intermediate fields, or compact representations to optimised material layouts. Early studies formulated TO as an image-to-image prediction problem using convolutional encoder--decoder networks~\cite{ulu2016data,sosnovik2019neural}. Subsequent works extended this paradigm to three-dimensional domains~\cite{banga20183d}, nonlinear structural problems~\cite{abueidda2020topology}, MMC-based real-time topology prediction~\cite{zheng2021accurate}, operator- and transformer-based topology prediction models~\cite{liang2024fourier,lutheran2026physics}, and low-dimensional latent-space reconstruction of optimised topologies~\cite{garayalde2024real}. Related studies have also explored neural ordinary differential equation (ODE)-based representations to describe the continuous evolution of topology design variables during optimisation~\cite{ma2024node}. These surrogate-based methods can rapidly approximate optimised layouts, but they are usually designed to predict or reconstruct a single near-optimal topology for a given load and boundary conditions. They provide limited support for design exploration, where multiple feasible topology candidates are often needed to investigate alternative load-transfer patterns and design trade-offs~\cite{oh2019deep,jang2022generative,radler2025generative}. This limitation has motivated the development of generative TO methods, which aim to synthesise diverse topology candidates under prescribed physical conditions while maintaining structural performance and feasibility.

Generative TO provides a promising approach for rapid and diverse topology synthesis, but existing studies also show that engineering stability remains a central challenge~\cite{bai2025towards}. Early cGAN-based models, including the prediction--refinement framework of Yu et al.~\cite{yu2018deep} and Nie et al.~\cite{nie2021topologygan}, demonstrated the feasibility of conditional GAN for TO application, with physical-field inputs improving consistency with prescribed loading and boundary conditions. However, GAN-based topology generators may suffer from training instability and mode collapse, with limited generalisability to out-of-distribution (OOD) design conditions, and their generated samples can neglect performance-related requirements such as compliance and manufacturability~\cite{maze2023diffusion}. Diffusion-based generative TO models have recently been introduced to improve sample quality while better preserving structural performance. For example, TopoDiff uses surrogate models to predict properties such as compliance and manufacturability during generation, thereby guiding the denoising process towards structurally efficient and manufacturable designs~\cite{maze2023diffusion}. Latent-space diffusion further combines conditional latent diffusion with physics-related input fields and auxiliary penalties to improve compliance prediction, volume-fraction control, and structural connectivity~\cite{lutheran2025latent}. Giannone and Ahmed propose a generative optimisation framework that combines
diffusion-based topology generation with subsequent SIMP refinement. Their results
show that, without refinement, generated designs may contain floating material or
low-performance features under OOD constraint
settings~\cite{giannone2023diffusing}. These developments indicate that generative TO should not only produce diverse topology samples, but should also maintain structural performance, volume feasibility, connectivity, and consistency with prescribed physical conditions. Nevertheless, the improved stability of existing generative TO models is often achieved through external conditioning, surrogate-based guidance, auxiliary penalties, post-processing, or optimisation-based refinement, introducing additional modelling and computational overhead beyond the generative model itself~\cite{nie2021topologygan,maze2023diffusion,lutheran2025latent,yu2018deep,giannone2023diffusing}. In addition, diffusion-based methods usually require many iterative reverse-sampling steps, which can increase the cost of generating large numbers of candidate topologies~\cite{ho2020denoising,maze2023diffusion,lutheran2025latent}. These observations suggest that a suitable generative TO framework should avoid unstable training and reduce sampling cost while maintaining structural performance and physical consistency.

Flow matching has recently emerged as a promising alternative for addressing these challenges. Rather than relying on adversarial optimisation or iterative reverse denoising, it learns a continuous velocity field that transports samples from a simple source distribution to the target topology distribution through a supervised velocity-matching objective~\cite{lipman2022flow}. The resulting ordinary differential equation (ODE) can be solved efficiently during inference, making flow matching attractive for stable and efficient topology generation. In this work, linear FMTO is first established by directly interpolating between a Gaussian source field and the BESO reference topology under prescribed design conditions. The numerical results show that linear FMTO provides an efficient baseline, although its performance can remain unstable under challenging or limited-data conditions. Its intermediate states are direct mixtures of random noise and the final topology rather than mechanically meaningful topology-evolution states. These mechanically uninformative states can increase the difficulty of velocity-field regression and may leave high-compliance or physically unreasonable samples in the tail of the learned terminal distribution.

This issue is fundamentally associated with probability-path construction rather than merely the absence of additional conditioning variables. Flow matching does not prescribe a unique probability path, and different choices define different intermediate states and target velocities, thereby affecting velocity-field learning and ODE sampling stability~\cite{lipman2022flow,tong2023improving,liu2022flow,lim2024elucidating}. In linear FMTO, the path is defined solely by the source sample and final topology and therefore does not reflect the material redistribution governed by structural response, sensitivity information, and design constraints. The central question is therefore not only whether flow matching can generate topologies, but how a trajectory-aware probability path should be constructed to improve the quality and stability of topology generation while better reflecting the underlying optimisation process.

Conventional TO algorithms naturally generate intermediate design states that record how material is progressively redistributed during optimisation~\cite{bendsoe1989optimal,zhou1991coc,wang2003level,zhang2016new,xie1993simple,querin1998evolutionary}. Among these methods, BESO is particularly suitable for providing optimisation-history information because it evolves the topology through strain-energy- and sensitivity-driven material removal and addition, while its binary distribution offers a consistent topology that is easier for the model to learn~\cite{xie1993simple,querin1998evolutionary}. Its prescribed volume evolution also provides a consistent basis for indexing and aligning intermediate states across different optimisation cases. The resulting BESO history therefore provides a natural physics-guided basis for constructing the trajectory-aware path. However, the BESO trajectory is not imposed as an exact transport path because it represents one optimiser-specific evolution from a prescribed initial design, whereas flow matching transports different random source samples to the target topology distribution. Enforcing the recorded trajectory directly would impose the same optimisation history on distinct source--target transports. Instead, the BESO history guides probability-path construction during training, with a trajectory weight balancing this guidance against endpoint transport.

The main contribution of this work is a trajectory-aware FMTO framework that incorporates volume-fraction-indexed BESO states into both the probability path and the target velocity. Linear FMTO first establishes the endpoint-based flow matching framework and exposes the absence of mechanics-informed intermediate states in the standard path, which can limit topology-generation quality and stability. Trajectory-aware FMTO then addresses this limitation by introducing BESO optimisation history into path construction. A path--velocity mismatch analysis characterises the trade-off between endpoint transport and trajectory guidance and explains why moderate guidance may reduce the mismatch, whereas excessive guidance may over-constrain the learned transport. An error-propagation analysis further connects the path--velocity mismatch to velocity-field approximation error, terminal generation error, and the probability of generating high-compliance failure samples. The proposed framework and the trends predicted by the theoretical analysis are examined through numerical studies of linear FMTO, trajectory guidance under limited training data, trajectory weight, anchor density, sampling efficiency, and three-dimensional topology generation.

The remainder of this article is organised as follows. Section~\ref{sec2} reviews the minimum-compliance TO formulation, the BESO procedure, and the preliminaries of flow matching. Section~\ref{sec3} presents the proposed FMTO framework, including data preparation, linear and trajectory-aware path construction, path--velocity mismatch analysis, error propagation, sampling, and numerical implementation. Section~\ref{sec4} evaluates linear FMTO as the base generator and investigates the effects of trajectory guidance, trajectory weight, anchor density, limited training data, sampling efficiency, and extension to three-dimensional TO. Section~\ref{sec5} summarises the main findings, discusses the limitations, and outlines future research directions.

\section{Topology Optimisation and Flow Matching Preliminaries}\label{sec2}
This section outlines the fundamental formulations of the minimum-compliance TO problem and flow matching model, upon which the proposed FMTO methodology is constructed.

\subsection{Topology Optimisation}

TO seeks to determine the optimal layout of material within a prescribed design domain while satisfying the governing physical equations and design constraints. In structural compliance minimisation, the objective is to obtain a stiff load-bearing structure under a prescribed material-volume constraint~\cite{bendsoe2013topology}. In this study, the soft-kill BESO method is adopted to generate the reference topologies for the learning framework~\cite{huang2007convergent}.

The design domain is discretised into $N$ finite elements. Each element is assigned a binary material density variable $\rho_i$, where $\rho_i=1$ denotes solid material and $\rho_i=\rho_{\min}$ denotes void material with a small residual stiffness to avoid singularity. The minimum-compliance problem is written as

\begin{equation}
\begin{aligned}
\boldsymbol{\rho}^{*}
=
\underset{\boldsymbol{\rho}}{\operatorname{arg\,min}}
\quad
&
C(\boldsymbol{\rho})
=
\boldsymbol{F}^{\mathrm{T}}\boldsymbol{U}
=
\boldsymbol{U}^{\mathrm{T}}
\boldsymbol{K}(\boldsymbol{\rho})
\boldsymbol{U},
\\
\text{s.t.}\quad
&
\boldsymbol{K}(\boldsymbol{\rho})\boldsymbol{U}
=
\boldsymbol{F},
\\
&
\frac{\sum_{i=1}^{N}\rho_i v_i}{\sum_{i=1}^{N}v_i}
\leq V_f,
\\
&
\rho_i\in\{\rho_{\min},1\},
\qquad i=1,\ldots,N.
\end{aligned}
\label{eq:topology_optimisation}
\end{equation}

where $\boldsymbol{\rho}=[\rho_1,\rho_2,\ldots,\rho_N]^{\mathrm{T}}$ denotes the material distribution, $\boldsymbol{K}(\boldsymbol{\rho})$ is the global stiffness matrix, and $\boldsymbol{U}$ and $\boldsymbol{F}$ are the global displacement and load vectors, respectively. In addition, $v_i$ is the volume of the $i$th element, $V_f$ is the prescribed volume-fraction, and $\rho_{\min}$ is a small positive density assigned to void elements.

In BESO, the topology is evolved by addition and removal of elements according to their sensitivity numbers. After solving the finite element equilibrium equation, the sensitivity number of the $i$th element for compliance minimisation is evaluated from the elemental strain energy as

\begin{equation}
\alpha_i
=
-\frac{\partial C}{\partial \rho_i}
\propto
\boldsymbol{u}_i^{\mathrm{T}}
\boldsymbol{k}_i^{0}
\boldsymbol{u}_i,
\label{eq:beso_sensitivity}
\end{equation}

where $\boldsymbol{u}_i$ is the elemental displacement vector and $\boldsymbol{k}_i^{0}$ is the elemental stiffness matrix of solid material. The sensitivity numbers are commonly filtered over neighbouring elements to reduce checkerboard patterns and mesh-dependent features,

\begin{equation}
\tilde{\alpha}_i
=
\frac{
\sum_{j\in\mathcal{N}_i} w_{ij}\alpha_j
}{
\sum_{j\in\mathcal{N}_i} w_{ij}
},
\qquad
w_{ij}
=
\max(0,r_{\min}-d_{ij}),
\label{eq:beso_filter}
\end{equation}

where $\mathcal{N}_i$ denotes the neighbourhood of element $i$, $d_{ij}$ is the distance between elements $i$ and $j$, and $r_{\min}$ is the filter radius. At each iteration, the intermediate volume target is first updated according to the evolutionary ratio,

\begin{equation}
V^{k}
=
\max\left((1-e_r)V^{k-1},V_f\right),
\label{eq:beso_volume_evolution}
\end{equation}

where $e_r$ is the evolutionary ratio and $V^{k}$ is the volume target at the $k$th iteration. After finite element analysis, the raw sensitivity numbers are spatially filtered and then averaged with the sensitivity numbers from the previous iteration to stabilise the evolutionary process. The elements are subsequently ranked according to the stabilised sensitivity numbers under the current volume target. Low-sensitivity solid elements are removed, while high-sensitivity void elements may be reintroduced as solid material. This bi-directional add--remove process allows the topology to evolve while following the prescribed volume reduction schedule. The binary topology is updated iteratively until the prescribed volume-fraction is reached and the relative change in the objective value falls below a predefined convergence tolerance.

\subsection{Flow Matching}

Flow matching is a generative modelling framework that learns a time-dependent neural velocity field to continuously transform samples from a simple source distribution to a target data distribution~\cite{lipman2022flow,tong2023improving}. Let $p_0(\boldsymbol{x})$ and $p_1(\boldsymbol{x})$ denote the source and target distributions, respectively. Flow matching constructs a continuous probability path $\{p_t\}_{t\in[0,1]}$ satisfying

\begin{equation}
p_{t=0}=p_0,
\qquad
p_{t=1}=p_1.
\label{eq:fm_boundary_conditions}
\end{equation}

The evolution of a sample along this path is governed by the ordinary differential equation

\begin{equation}
\frac{\mathrm{d}\boldsymbol{x}_t}{\mathrm{d}t}
=
\boldsymbol{v}_{\theta}(\boldsymbol{x}_t,t),
\qquad
t\in[0,1],
\label{eq:fm_ode}
\end{equation}

where $\boldsymbol{x}_t$ denotes the intermediate state at time $t$, and $\boldsymbol{v}_{\theta}$ is a time-dependent neural velocity field. Starting from $\boldsymbol{x}_0\sim p_0$, integrating Eq.~\eqref{eq:fm_ode} from $t=0$ to $t=1$ yields a terminal sample whose distribution approximates $p_1$. During training, flow matching directly matches the target velocity associated with a prescribed probability path and therefore avoids numerical integration of the generative ODE.

Given a probability path $\{p_t\}_{t\in[0,1]}$ connecting $p_0$ and $p_1$, let $\boldsymbol{u}_t(\boldsymbol{x})$ denote the corresponding marginal velocity field. In principle, the neural velocity field can be trained using the flow matching objective

\begin{equation}
\mathcal{L}_{\mathrm{FM}}(\theta)
=
\mathbb{E}_{t\sim\mathcal{U}(0,1),\,
\boldsymbol{x}_t\sim p_t}
\left[
\left\|
\boldsymbol{v}_{\theta}(\boldsymbol{x}_t,t)
-
\boldsymbol{u}_t(\boldsymbol{x}_t)
\right\|_2^2
\right].
\label{eq:fm_loss}
\end{equation}

\noindent However, the marginal velocity field $\boldsymbol{u}_t(\boldsymbol{x})$ is generally unavailable in closed form. Conditional flow matching therefore constructs tractable conditional probability paths associated with sampled source--target endpoints $\boldsymbol{x}_0\sim p_0$ and $\boldsymbol{x}_1\sim p_1$. For a randomly sampled time $t\sim\mathcal{U}(0,1)$, an intermediate state $\boldsymbol{x}_t$ is drawn from the corresponding conditional path $p_t(\boldsymbol{x}\mid\boldsymbol{x}_0,\boldsymbol{x}_1)$. The neural velocity field is then trained using

\begin{equation}
\mathcal{L}_{\mathrm{CFM}}(\theta)
=
\mathbb{E}
\left[
\left\|
\boldsymbol{v}_{\theta}(\boldsymbol{x}_t,t)
-
\boldsymbol{u}_t
(\boldsymbol{x}_t\mid\boldsymbol{x}_0,\boldsymbol{x}_1)
\right\|_2^2
\right],
\label{eq:cfm_loss}
\end{equation}

\noindent where the expectation is taken over the sampled source--target pairs, flow times, and intermediate states.

The conditional flow matching formulation does not prescribe a unique conditional probability path. Different path constructions may be adopted, provided that the intermediate states and their corresponding target velocities can be evaluated during training~\cite{lipman2024flow}. The selected path determines the intermediate states presented to the velocity network and can influence the geometry of the learned flow, training behaviour, and numerical sampling efficiency~\cite{lim2024elucidating}.

For topology generation, the target sample $\boldsymbol{x}_1$ corresponds to an optimised topology field, while the velocity network is additionally conditioned on the physical and design variables $\boldsymbol{c}$. The resulting problem-conditioned velocity field is denoted by
$\boldsymbol{v}_{\theta}(\boldsymbol{x}_t,t,\boldsymbol{c})$. Based on this general formulation, the linear and BESO-guided trajectory-aware paths considered in this study are introduced in the following section.

\section{Trajectory-aware Flow Matching Topology Optimisation
Framework}\label{sec3}

This section presents the proposed FMTO framework with an emphasis on trajectory-aware path
construction. The purpose is not only to define a conditional generative model for topology
fields, but also to show how optimisation-history information can be converted into a training-time
probability path for flow matching.

\subsection{Data preparation and BESO trajectory recording}
\label{subsec:data_preparation}

For a given set of physical and design conditions, the BESO produces a converged
numerical reference topology,

\begin{equation}
\boldsymbol{\rho}^{\mathrm{ref}}_i
=
\mathcal{T}_{\mathrm{BESO}}(\boldsymbol{c}_i),
\label{eq:to_mapping}
\end{equation}

\noindent where \(\boldsymbol{c}_i\) denotes the conditions of the \(i\)th TO instance and
\(\boldsymbol{\rho}^{\mathrm{ref}}_i\) is the corresponding BESO solution. The notation
\(\boldsymbol{\rho}^{*}\) is reserved for the ideal optimum of the mathematical TO problem in
Eq.~\eqref{eq:topology_optimisation}. For linear FMTO, the training data consist of
condition--topology pairs,

\begin{equation}
\mathcal{D}_{\mathrm{lin}}
=
\left\{
\left(
\boldsymbol{c}_i,
\boldsymbol{\rho}^{\mathrm{ref}}_i
\right)
\right\}_{i=1}^{M},
\label{eq:linear_dataset}
\end{equation}

\noindent where \(M\) is the number of TO instances.

For trajectory-aware FMTO, representative intermediate BESO states are also recorded. These
states are saved at descending volume-fraction levels,

\begin{equation}
\mathcal{V}_i
=
\left\{
\nu_i^{(0)},
\nu_i^{(1)},
\ldots,
\nu_i^{(L)}
\right\},
\qquad
\nu_i^{(0)}
>
\nu_i^{(1)}
>
\cdots
>
\nu_i^{(L)},
\label{eq:trajectory_volume_fraction_levels}
\end{equation}

\noindent with the corresponding topology trajectory

\begin{equation}
\mathcal{R}_i
=
\left\{
\boldsymbol{\rho}_i^{(0)},
\boldsymbol{\rho}_i^{(1)},
\ldots,
\boldsymbol{\rho}_i^{(L)}
\right\},
\qquad
\boldsymbol{\rho}_i^{(L)}
=
\boldsymbol{\rho}^{\mathrm{ref}}_i.
\label{eq:beso_trajectory}
\end{equation}

\noindent The trajectory dataset is therefore

\begin{equation}
\mathcal{D}_{\mathrm{traj}}
=
\left\{
\left(
\boldsymbol{c}_i,
\mathcal{R}_i,
\mathcal{V}_i
\right)
\right\}_{i=1}^{M}.
\label{eq:trajectory_dataset}
\end{equation}

Volume-fraction is used to align optimisation progress because different TO instances may converge
in different numbers of iterations. The recorded states are mapped to the flow time \(t\in[0,1]\)
by

\begin{equation}
t_{i,\ell}
=
\frac{
\nu_i^{(0)}-\nu_i^{(\ell)}
}{
\nu_i^{(0)}-\nu_i^{(L)}
},
\qquad
\ell=0,\ldots,L.
\label{eq:vf_to_flow_time}
\end{equation}

\noindent Thus, \(t_{i,0}=0\) corresponds to the initial material distribution and \(t_{i,L}=1\)
corresponds to the converged reference topology. The BESO trajectory is not treated as an
additional final target. Instead, it is used to define a training-time guide for the probability
path.

The condition \(\boldsymbol{c}_i\) contains global descriptors and spatial fields encoding the
prescribed design variables, supports, and loads. Gaussian source fields are sampled online as

\begin{equation}
\boldsymbol{x}_0
\sim
p_0(\boldsymbol{x})
=
\mathcal{N}(\boldsymbol{0},\boldsymbol{I}),
\label{eq:fm_source_sampling}
\end{equation}

\noindent with the same spatial resolution as the topology field. For linear FMTO,
\(\boldsymbol{x}_0\) is paired with \(\boldsymbol{\rho}^{\mathrm{ref}}_i\). For
trajectory-aware FMTO, \(\boldsymbol{x}_0\) is paired with \(\mathcal{R}_i\) and
\(\mathcal{V}_i\).

\label{subsec:FMTO_architecture}

\begin{figure}[htbp]
    \centering
    \includegraphics[width=0.93\textwidth]{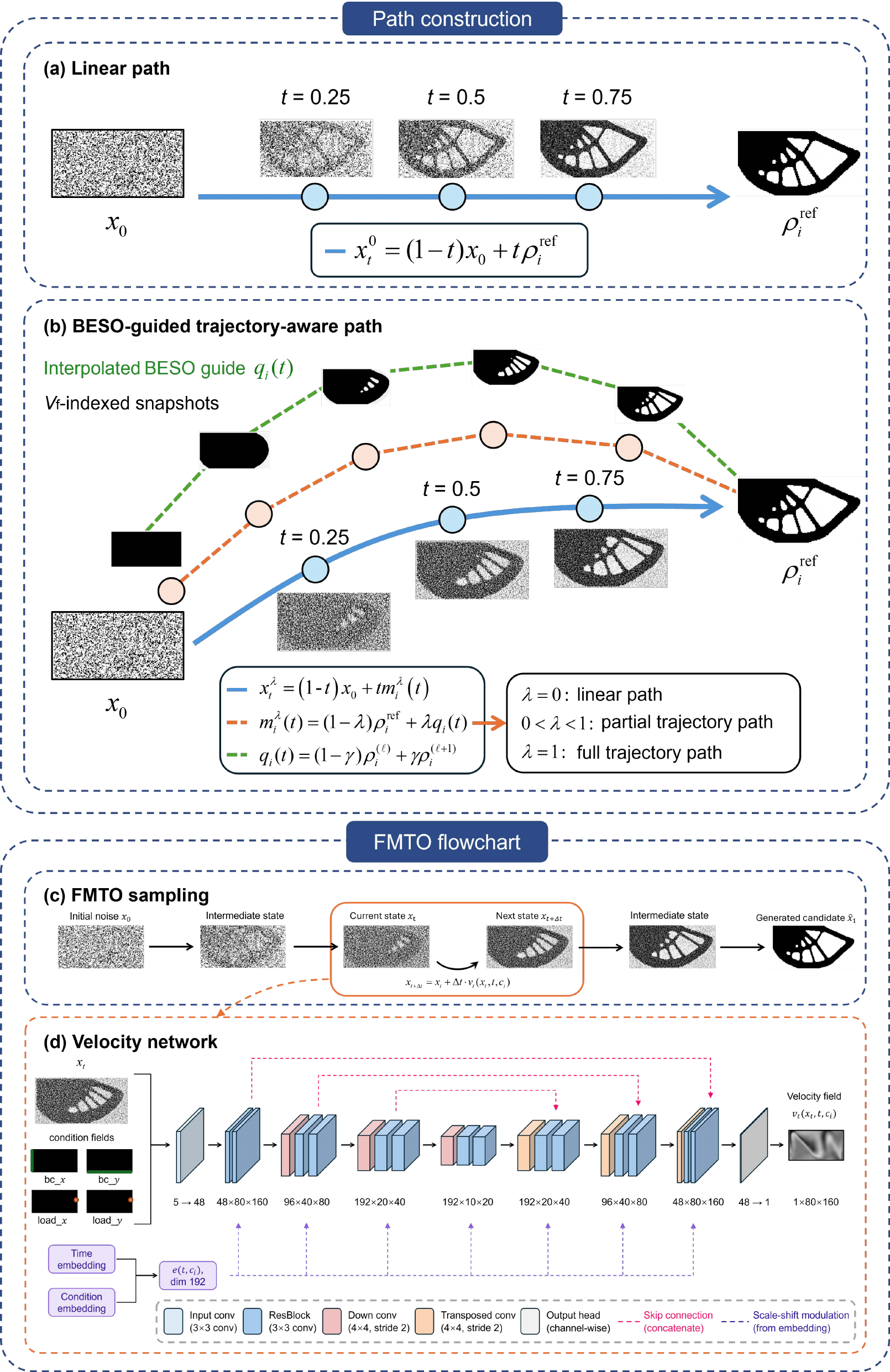}
    \caption{Schematic illustration of the proposed FMTO path construction and sampling
workflow. (a) Endpoint-based linear path from a Gaussian source field to the BESO reference
topology. (b) BESO-guided trajectory-aware path, where the interpolated BESO guide
\(q_i(t)\) is used to construct the \(\lambda\)-weighted centreline \(m_i^\lambda(t)\) and the
corresponding intermediate state \(x_t^\lambda\). (c) ODE-based FMTO sampling from
\(x_0\) to the generated topology. (d) Conditional velocity network, which takes the current
state \(x_t\), flow time \(t\), and design condition \(c_i\) as inputs and outputs the velocity
field \(v_\theta(x_t,t,c_i)\).}
    \label{fig:fm_path}
\end{figure}

\subsection{Linear FMTO as an endpoint-based baseline}
\label{subsec:linear_fmto}

As illustrated in Fig.~\ref{fig:fm_path}(a), linear FMTO constructs an
endpoint-based straight-line probability path between a Gaussian source field and a
reference topology~\cite{albergo2025stochastic,liu2022flow}. For each training pair
\((\boldsymbol{c}_i,\boldsymbol{\rho}^{\mathrm{ref}}_i)
\in\mathcal{D}_{\mathrm{lin}}\), the intermediate state is defined as

\begin{equation}
\boldsymbol{x}^{0}_{t}
=
(1-t)\boldsymbol{x}_0
+
t\boldsymbol{\rho}^{\mathrm{ref}}_i,
\label{eq:linear_fm_path}
\end{equation}

\noindent where
\(t\sim\mathcal{U}(0,1)\) and
\(\boldsymbol{x}_0\sim\mathcal{N}(\boldsymbol{0},\boldsymbol{I})\).
The corresponding target velocity is

\begin{equation}
\boldsymbol{u}^{0}_{t}
=
\frac{\mathrm{d}\boldsymbol{x}^{0}_{t}}{\mathrm{d}t}
=
\boldsymbol{\rho}^{\mathrm{ref}}_i
-
\boldsymbol{x}_0.
\label{eq:linear_fm_velocity}
\end{equation}

\noindent Accordingly, the straight path shown in
Fig.~\ref{fig:fm_path}(a) evolves directly from the Gaussian source sample
\(\boldsymbol{x}_0\) towards the reference topology
\(\boldsymbol{\rho}^{\mathrm{ref}}_i\), without introducing any intermediate
optimisation-history information.

The conditional velocity network is trained to approximate the target velocity by
minimising

\begin{equation}
\mathcal{L}_{\mathrm{lin}}
=
\mathbb{E}_{
(\boldsymbol{c}_i,\boldsymbol{\rho}^{\mathrm{ref}}_i),
\boldsymbol{x}_0,t
}
\left[
\left\|
\boldsymbol{v}^{0}_{\theta}
\left(
\boldsymbol{x}^{0}_{t},
t,
\boldsymbol{c}_i
\right)
-
\boldsymbol{u}^{0}_{t}
\right\|_2^2
\right].
\label{eq:linear_fm_loss}
\end{equation}

\noindent The trained velocity field is subsequently used for generation through
the common ODE sampling procedure described in
Section~\ref{subsec:fmto_sampling}.

The linear formulation is simple and provides an efficient endpoint-based baseline.
However, as shown by the intermediate states along the path in
Fig.~\ref{fig:fm_path}(a), these states are direct mixtures of random noise and the
final reference topology. They do not represent mechanically meaningful stages of
topology evolution and therefore do not use the optimisation history through which
the reference design was obtained. Consequently, the network must regress the
velocity field from intermediate states that may be poorly aligned with physically
plausible material-redistribution patterns. This may increase the difficulty of
velocity-field learning and produce a broader terminal distribution, leaving
high-compliance or physically unreasonable samples in the generated tail. This
limitation motivates the BESO-guided trajectory-aware formulation illustrated in
Fig.~\ref{fig:fm_path}(b).

\subsection{BESO-guided trajectory-aware path}
\label{subsec:trajectory_fmto}

As illustrated in Fig.~\ref{fig:fm_path}(b), the trajectory-aware formulation uses
volume-fraction-indexed BESO states to construct a mechanics-informed guide for the
probability path. These states describe progressive material redistribution driven
by finite element equilibrium, strain-energy sensitivity, and volume evolution.
However, the BESO trajectory is not assumed to be an exact generative transport
path. Instead, it is introduced as a soft training-time guide for constructing
physically more informative intermediate states.

Let
\(\{\boldsymbol{\rho}^{(\ell)}_i\}_{\ell=0}^{L_i}\)
denote the recorded BESO states for the \(i\)-th design case, with the corresponding
normalised trajectory times
\(\{t_{i,\ell}\}_{\ell=0}^{L_i}\).
The discrete BESO states shown along the green trajectory in
Fig.~\ref{fig:fm_path}(b) are converted into a continuous trajectory guide
\(\boldsymbol{q}_i(t)\) through piecewise-linear interpolation. For
\(t\in[t_{i,\ell},t_{i,\ell+1}]\),

\begin{equation}
\boldsymbol{q}_i(t)
=
(1-\gamma)
\boldsymbol{\rho}^{(\ell)}_i
+
\gamma
\boldsymbol{\rho}^{(\ell+1)}_i,
\qquad
\gamma
=
\frac{
t-t_{i,\ell}
}{
t_{i,\ell+1}-t_{i,\ell}
}.
\label{eq:trajectory_interpolation}
\end{equation}

\noindent The corresponding segment-wise trajectory velocity is

\begin{equation}
\dot{\boldsymbol{q}}_i(t)
=
\frac{
\boldsymbol{\rho}^{(\ell+1)}_i
-
\boldsymbol{\rho}^{(\ell)}_i
}{
t_{i,\ell+1}-t_{i,\ell}
}.
\label{eq:trajectory_derivative}
\end{equation}

\noindent The piecewise-linear trajectory
\(\boldsymbol{q}_i(t)\) is continuous, although its derivative may be
discontinuous at the interpolation knots. Since these knots form a set of measure
zero under continuous sampling of \(t\), they do not affect the expected
flow-matching objective.

The interpolated BESO trajectory is not imposed directly as the probability-path
centreline. Instead, as represented by the orange trajectory in
Fig.~\ref{fig:fm_path}(b), a trajectory weight
\(\lambda\in[0,1]\) is introduced to balance endpoint supervision and
optimisation-history guidance. The guided centreline is defined as

\begin{equation}
\boldsymbol{m}^{\lambda}_i(t)
=
(1-\lambda)
\boldsymbol{\rho}^{\mathrm{ref}}_i
+
\lambda
\boldsymbol{q}_i(t),
\label{eq:guided_center}
\end{equation}

\noindent with the corresponding centreline velocity

\begin{equation}
\dot{\boldsymbol{m}}^{\lambda}_i(t)
=
\lambda
\dot{\boldsymbol{q}}_i(t).
\label{eq:guided_center_velocity}
\end{equation}

\noindent Here, \(\lambda=0\) recovers the endpoint-based centreline used by the
linear formulation in Fig.~\ref{fig:fm_path}(a), whereas \(\lambda=1\)
makes the centreline fully follow the interpolated BESO guide
\(\boldsymbol{q}_i(t)\). Intermediate values of \(\lambda\) produce the
orange \(\lambda\)-weighted centreline in Fig.~\ref{fig:fm_path}(b), thereby
providing a soft balance between supervision from the final reference topology and
guidance from the BESO optimisation trajectory.

Using this time-dependent centreline, the trajectory-aware probability path shown
by the blue curve in Fig.~\ref{fig:fm_path}(b) is defined as

\begin{equation}
\boldsymbol{x}^{\lambda}_{t}
=
(1-t)\boldsymbol{x}_0
+
t\boldsymbol{m}^{\lambda}_i(t).
\label{eq:trajectory_path}
\end{equation}

\noindent Unlike the endpoint-based path in
Eq.~\eqref{eq:linear_fm_path}, this path is not a straight-line interpolation
because its centreline
\(\boldsymbol{m}^{\lambda}_i(t)\) varies with time. Differentiating
Eq.~\eqref{eq:trajectory_path} gives the corresponding target velocity

\begin{equation}
\boldsymbol{u}^{\lambda}_{t}
=
\frac{\mathrm{d}\boldsymbol{x}^{\lambda}_{t}}{\mathrm{d}t}
=
-
\boldsymbol{x}_0
+
\boldsymbol{m}^{\lambda}_i(t)
+
t\dot{\boldsymbol{m}}^{\lambda}_i(t).
\label{eq:trajectory_velocity}
\end{equation}

\noindent The final term
\(t\dot{\boldsymbol{m}}^{\lambda}_i(t)\)
accounts explicitly for the time variation of the trajectory-guided centreline and
distinguishes the proposed target velocity from that of the endpoint-based linear
path.

The trajectory-aware conditional velocity network is trained by minimising

\begin{equation}
\mathcal{L}_{\mathrm{traj}}
=
\mathbb{E}_{
(\boldsymbol{c}_i,\mathcal{R}_i,\mathcal{V}_i),
\boldsymbol{x}_0,t
}
\left[
\left\|
\boldsymbol{v}^{\lambda}_{\theta}
\left(
\boldsymbol{x}^{\lambda}_{t},
t,
\boldsymbol{c}_i
\right)
-
\boldsymbol{u}^{\lambda}_{t}
\right\|_2^2
\right],
\label{eq:trajectory_fm_loss}
\end{equation}

\noindent where \(\mathcal{R}_i\) denotes the recorded BESO topology states and
\(\mathcal{V}_i\) denotes their associated volume-fraction sequence.

The complete trajectory-aware path construction is summarised in
Fig.~\ref{fig:fm_path}(b). Importantly, the BESO trajectory is required only during
training to construct
\(\boldsymbol{x}^{\lambda}_{t}\) and
\(\boldsymbol{u}^{\lambda}_{t}\). During inference, neither the recorded BESO states
nor the interpolated guide \(\boldsymbol{q}_i(t)\) is required. Linear and
trajectory-aware FMTO therefore follow the same ODE sampling procedure described in
Section~\ref{subsec:fmto_sampling}.

\subsection{Theoretical rationale for trajectory guidance}
\label{subsec:path_velocity_analysis}

The benefit of trajectory guidance can be understood from a path-approximation perspective.
Flow matching learns a velocity field, and different paths expose the network to different
intermediate states and target velocities. A trajectory guide is useful only if it reduces the
mismatch between the constructed training path and a more favourable, but unknown, generative path
for topology evolution.

Let \(\boldsymbol{m}^{*}_i(t)\) denote a hypothetical ideal generative centreline. The detailed
derivation is provided in~\ref{app:path_velocity_mismatch}. In compact form, the
combined path--velocity mismatch of the trajectory-aware centreline can be written as

\begin{equation}
G(\lambda)
=
(1-\lambda)^2B_G
+
\lambda^2N_G
+
2\lambda(1-\lambda)C_G.
\label{eq:path_velocity_mismatch}
\end{equation}

\noindent Here, \(B_G\) measures the mismatch of the endpoint-based linear path, \(N_G\) measures
the
mismatch of the BESO trajectory guide, and \(C_G\) measures their correlation. Because
\(G(0)=B_G\), trajectory guidance improves the path design when \(G(\lambda)<G(0)\).
The unconstrained minimiser of this quadratic mismatch is

\begin{equation}
\lambda^{*}
=
\frac{
B_G-C_G
}{
B_G+N_G-2C_G
},
\label{eq:lambda_star}
\end{equation}

\noindent with the practical value constrained to \([0,1]\). This expression shows that increasing
\(\lambda\) does not necessarily improve the path. Instead, a moderate value may better balance
endpoint supervision and BESO trajectory guidance, whereas excessive guidance may over-constrain the
learned transport.

The same interpretation can be connected to generation stability. Under finite data and finite
model capacity, reducing \(G(\lambda)\) can reduce the difficulty of velocity-field regression.
Through the stability of the generative ODE, a smaller velocity-field error can in turn reduce the
terminal generation error and the upper bound on the probability of generating high-compliance
samples. The full error-propagation argument and its assumptions are given in~\ref{app:error_propagation}. This analysis is used as an interpretation rather than an
unconditional guarantee.

\subsection{Sampling, post-processing, and FEA evaluation}
\label{subsec:fmto_sampling}

After training, the linear and trajectory-aware FMTO models use the same inference
procedure. As illustrated by the global sampling sequence in
Fig.~\ref{fig:fm_path}(c), sampling starts from a Gaussian source field and advances
the topology state from \(t=0\) to \(t=1\). For a prescribed design condition
\(\boldsymbol{c}\), the initial state is sampled according to
Eq.~\eqref{eq:fm_source_sampling}, and the learned ordinary differential equation is

\begin{equation}
\frac{\mathrm{d}\boldsymbol{x}_t}{\mathrm{d}t}
=
\boldsymbol{v}_{\theta}
\left(
\boldsymbol{x}_t,
t,
\boldsymbol{c}
\right),
\qquad
\boldsymbol{x}_{t=0}
=
\boldsymbol{x}_0.
\label{eq:fmto_sampling_ode}
\end{equation}

\noindent At each solver evaluation, the current topology state
\(\boldsymbol{x}_t\), the current flow time \(t\), and the prescribed condition
\(\boldsymbol{c}\) are passed to the conditional U-Net, as illustrated in
Fig.~\ref{fig:fm_path}(d). The network predicts the instantaneous velocity field

\begin{equation}
\widehat{\boldsymbol{u}}_t
=
\boldsymbol{v}_{\theta}
\left(
\boldsymbol{x}_t,
t,
\boldsymbol{c}
\right).
\label{eq:predicted_sampling_velocity}
\end{equation}

\noindent The predicted velocity is then used by the numerical ODE solver to advance
the topology state. For example, an explicit Euler update is written as

\begin{equation}
\boldsymbol{x}_{t_{k+1}}
=
\boldsymbol{x}_{t_k}
+
\Delta t\,
\boldsymbol{v}_{\theta}
\left(
\boldsymbol{x}_{t_k},
t_k,
\boldsymbol{c}
\right),
\qquad
t_{k+1}=t_k+\Delta t.
\label{eq:fmto_euler_update}
\end{equation}

\noindent Repeated velocity-network evaluations and ODE updates produce the
sampling sequence

\begin{equation}
\boldsymbol{x}_0
\longrightarrow
\boldsymbol{x}_{t_1}
\longrightarrow
\boldsymbol{x}_{t_2}
\longrightarrow
\cdots
\longrightarrow
\widehat{\boldsymbol{x}}_1,
\label{eq:fmto_sampling_sequence}
\end{equation}

\noindent where \(\widehat{\boldsymbol{x}}_1\) is the generated continuous topology
field at \(t=1\). Multiple candidates for the same condition are obtained by drawing
different Gaussian source fields \(\boldsymbol{x}_0\).

The two FMTO formulations differ only in their training-time path construction and
velocity-field supervision. Linear FMTO is trained using
Eq.~\eqref{eq:linear_fm_loss}, whereas trajectory-aware FMTO is trained using
Eq.~\eqref{eq:trajectory_fm_loss}. The recorded BESO states, the interpolated guide
\(\boldsymbol{q}_i(t)\), and the guided centreline
\(\boldsymbol{m}^{\lambda}_i(t)\) are required only during training and are not
evaluated during inference. Therefore, trajectory guidance introduces no additional
BESO optimisation or trajectory-construction step into the sampling procedure.

For post-processing, each element of the generated terminal field is first clamped
to the interval \([0,1]\),

\begin{equation}
\widetilde{x}_{1,e}
=
\min
\left[
1,
\max
\left(
0,
\widehat{x}_{1,e}
\right)
\right],
\label{eq:fmto_clamping}
\end{equation}

\noindent and subsequently binarised using a fixed threshold of 0.5,

\begin{equation}
\widehat{\rho}^{\mathrm{bin}}_e
=
\begin{cases}
1,
&
\widetilde{x}_{1,e}>0.5,
\\[2pt]
0,
&
\widetilde{x}_{1,e}\leq 0.5.
\end{cases}
\label{eq:fmto_binarisation}
\end{equation}

\noindent Here, \(e\) denotes the element index. Elements assigned a value of one
are treated as solid, whereas the remaining elements are treated as void. The same
thresholded topology is used for visualisation and finite element evaluation.

Each generated topology is re-evaluated using the same boundary conditions, loads,
material properties, mesh, and finite element settings as the corresponding BESO
reference. When multiple candidates are generated for the same condition, their
FEA-based evaluation and ranking are performed only after neural-network sampling
and do not alter the generative ODE. The quantitative measures used to assess
structural performance, volume feasibility, and topology fidelity are introduced in
the following subsection.

\subsection{Evaluation metrics}
\label{subsec:evaluation_metrics}

Following the sampling, post-processing, and FEA procedure described in
Section~\ref{subsec:fmto_sampling}, the generated samples are evaluated in terms of
structural performance, volume feasibility, and topology fidelity. Compliance- and
volume-related metrics are computed from the thresholded binary topologies, whereas
Raw MAE is computed from the clamped continuous density fields before thresholding.

For the \(j\)-th generated sample of validation case \(i\), 
\(C/C_{\mathrm{BESO}}\) denotes the ratio between the recomputed compliance of
the generated topology, \(C_{\mathrm{gen},i}^{(j)}\), and that of the corresponding
BESO reference, \(C_{\mathrm{BESO},i}\). A value below one indicates lower
recomputed compliance than the reference under the adopted FEA settings.

The median \(C/C_{\mathrm{BESO}}\) is evaluated over all generated samples. The
mean-best \(C/C_{\mathrm{BESO}}\) is obtained by selecting the lowest ratio for
each validation case and then averaging across cases. The constrained mean-best
value is computed in the same manner after retaining only samples satisfying
\(V_{f,i}^{(j)}\leq V_{f,i}^{*}\). The failure rate is defined as the percentage of generated samples with
\(C/C_{\mathrm{BESO}}>1.25\).

The generated volume-fraction is

\begin{equation}
V_{f,i}^{(j)}
=
\frac{
\sum_{e=1}^{N_e}
\widehat{\rho}_{i,e}^{(j),\mathrm{bin}}v_e
}{
\sum_{e=1}^{N_e}v_e
},
\label{eq:metric_volume_fraction}
\end{equation}

\noindent where \(v_e\) is the volume of element \(e\). The feasible-sample rate is
the percentage of samples satisfying
\(V_{f,i}^{(j)}\leq V_{f,i}^{*}\), while the mean absolute volume-fraction error is

\begin{equation}
E_{V_f}
=
\frac{1}{N_cN_s}
\sum_{i=1}^{N_c}
\sum_{j=1}^{N_s}
\left|
V_{f,i}^{(j)}
-
V_{f,i}^{*}
\right|,
\label{eq:metric_volume_error}
\end{equation}

\noindent where \(N_c\) and \(N_s\) are the numbers of validation cases and
generated samples per case, respectively.

Topology fidelity is evaluated relative to the corresponding BESO reference. Let
\(\Omega_i^{(j)}\) and \(\Omega_i^{\mathrm{ref}}\) denote the solid-element sets of
the thresholded generated and reference topologies. The intersection over union
(IoU) and Dice coefficient are defined as

\begin{equation}
\operatorname{IoU}_{i}^{(j)}
=
\frac{
\left|
\Omega_i^{(j)}
\cap
\Omega_i^{\mathrm{ref}}
\right|
}{
\left|
\Omega_i^{(j)}
\cup
\Omega_i^{\mathrm{ref}}
\right|
},
\qquad
\operatorname{Dice}_{i}^{(j)}
=
\frac{
2
\left|
\Omega_i^{(j)}
\cap
\Omega_i^{\mathrm{ref}}
\right|
}{
\left|
\Omega_i^{(j)}
\right|
+
\left|
\Omega_i^{\mathrm{ref}}
\right|
}.
\label{eq:metric_iou_dice}
\end{equation}

\noindent Boundary F1 measures agreement between the generated and reference
material boundaries and is defined as

\begin{equation}
\operatorname{BF1}
=
\frac{
2P_{\mathrm{B}}R_{\mathrm{B}}
}{
P_{\mathrm{B}}+R_{\mathrm{B}}
},
\label{eq:metric_boundary_f1}
\end{equation}

\noindent where \(P_{\mathrm{B}}\) and \(R_{\mathrm{B}}\) are the boundary precision
and recall evaluated using a one-pixel matching tolerance.

Raw MAE measures the element-wise difference between the clamped continuous
generated density field and the BESO reference,

\begin{equation}
\operatorname{MAE}_{i}^{(j)}
=
\frac{1}{N_e}
\sum_{e=1}^{N_e}
\left|
\widetilde{x}_{1,i,e}^{(j)}
-
\rho_{i,e}^{\mathrm{ref}}
\right|.
\label{eq:metric_raw_mae}
\end{equation}

\noindent The reported topology-fidelity metrics are averaged over all validation
cases and generated samples. Higher values are preferred for IoU, Dice, Boundary F1,
and feasible-sample rate, whereas lower values are preferred for compliance ratio,
failure rate, volume-fraction error, and Raw MAE.

\subsection{Network and implementation details}
\label{subsec:numerical_implementation}

As illustrated in Fig.~\ref{fig:fm_path}(d), the conditional velocity field
\(\boldsymbol{v}_{\theta}(\boldsymbol{x}_t,t,\boldsymbol{c})\) is represented by a
conditional U-Net. For the two-dimensional model, the current topology field
\(\boldsymbol{x}_t\) is concatenated with four spatial fields encoding the boundary
conditions and loads, forming a \(5\times80\times160\) input tensor. An initial
\(3\times3\) convolution maps the input to 48 channels. The encoder uses residual
blocks with feature channels \(48\), \(96\), and \(192\), followed by a
\(192\times10\times20\) bottleneck. The decoder mirrors the encoder through
transposed-convolution upsampling and skip-feature concatenation, and the output head
maps the final 48-channel feature field to a one-channel velocity field. Group
normalisation and SiLU activation are used throughout.

For the three-dimensional experiments, the input has a spatial resolution of
\(48\times24\times24\), and the same channel progression is retained. All
two-dimensional convolution, downsampling, and transposed-convolution operations are
replaced by their three-dimensional counterparts, resulting in a
\(128\times6\times3\times3\) bottleneck and a one-channel three-dimensional velocity
field.

The flow time is represented using sinusoidal features, while the global condition
vector is mapped through a multilayer perceptron. These representations are combined
into a 192-dimensional embedding
\(\boldsymbol{e}(t,\boldsymbol{c})\), which modulates each residual block through
scale--shift conditioning. The same conditional architecture is used for linear and
trajectory-aware FMTO. Within each dimensional setting, the DMTO baseline, linear
FMTO, and trajectory-aware FMTO are configured with the same number of trainable
parameters, ensuring that the comparisons are not affected by differences in model
capacity.

For each mini-batch, Gaussian source fields are sampled online from
\(\mathcal{N}(\boldsymbol{0},\boldsymbol{I})\), and flow times are sampled uniformly
from \([0,1]\). The intermediate states and target velocities are computed from the
corresponding path formulation, and the model is trained using the mean-squared
velocity-matching loss. AdamW is used with a learning rate of \(2\times10^{-4}\),
weight decay of \(1\times10^{-6}\), batch size of 8, gradient clipping with a
maximum norm of 1.0, and 500 training epochs. During inference, samples are generated
using explicit Euler integration. All computations were performed on an Intel Core
i9-13900K CPU, an NVIDIA GeForce RTX 4080 GPU, and 32~GB RAM.

\section{Numerical Examples}\label{sec4}

As outlined in the Introduction, the numerical studies are designed to examine three
questions arising from the proposed trajectory-aware FMTO framework:

\begin{itemize}
    \item Can endpoint-based linear FMTO serve as an efficient conditional generator
    for topology optimisation compared with a diffusion-based generative baseline?

    \item Can BESO optimisation histories improve flow matching by providing a
    trajectory-aware probability path, particularly when the available training data
    are limited?

    \item How sensitive is the proposed path construction to the discretisation of the
    BESO trajectory, and can the same principle be extended to three-dimensional
    topology generation with objective assessment of voxel-level topology fidelity and volume feasibility?
\end{itemize}

The numerical results are organised into four examples.
\hyperref[Basic_Linear_FMTO]{Example~1} establishes linear FMTO as an
endpoint-based baseline for two-dimensional conditional topology generation and
compares its generation quality and sampling efficiency with those of a
diffusion-based baseline.
\hyperref[Trajectory_Aware_FMTO]{Example~2} evaluates the proposed
trajectory-aware path under a limited-data setting and examines the influence of the
trajectory weight \(\lambda\), thereby testing whether moderate BESO trajectory
guidance improves generation quality relative to the linear path.
\hyperref[Trajectory_Anchor_Density]{Example~3} investigates the effect of
trajectory-anchor density, which controls the discretisation of the recorded BESO
optimisation history used for path construction.
Finally, \hyperref[ThreeD_FMTO]{Example~4} extends the evaluation to
three-dimensional topology generation and examines whether the proposed framework
can preserve structural validity beyond the two-dimensional setting.

Unless otherwise stated, \hyperref[Basic_Linear_FMTO]{Examples~1},
\hyperref[Trajectory_Aware_FMTO]{2}, and
\hyperref[Trajectory_Anchor_Density]{3} are conducted using two-dimensional BESO
datasets with a topology resolution of \(80\times160\). This common configuration
allows the endpoint-based baseline, trajectory-aware guidance, and trajectory
discretisation to be evaluated without additional variation from mesh resolution or
problem dimensionality. The three-dimensional extension is examined separately in
\hyperref[ThreeD_FMTO]{Example~4} using the corresponding voxel-based dataset and
model with a resolution of \(48\times24\times24\).

\subsection{Example 1: Linear FMTO as an Efficient Endpoint-Based Baseline}\label{Basic_Linear_FMTO}

This example is designed to validate the basic performance of the linear FMTO model in generating two-dimensional topology fields under prescribed physical and design conditions. As the first validation case, it provides a comprehensive comparison with the BESO reference solutions and a diffusion model baseline, focusing on the qualitative comparison of representative cases, overall generation quality and computational efficiency of the proposed framework.

In this example, a dataset of 5000 BESO-generated TO instances is used and divided into training and testing sets with a ratio of 9:1. Each instance contains the converged BESO reference topology, the global condition vector, and the spatial boundary-condition and load fields. The material properties are specified in non-dimensional form as solid Young's modulus \(E_0=1\) and Poisson's ratio \(\nu=0.3\). The target volume-fraction is set to \(V_f=0.5\), and the minimum stiffness for void elements in finite-element compliance evaluation is \(E_{\min}=1\times10^{-9}E_0\). The TO parameters are chosen as filter radius \(r_{\min}=5\), evolutionary rate \(er=0.01\), and convergence tolerance \(\tau=10^{-3}\). The default sampling settings are 20 Euler steps for FMTO and 1000 reverse-diffusion steps for DMTO. These settings are further examined through the step-sensitivity ablation at the end of this subsection.

\begin{figure}[htbp]
    \centering
    \includegraphics[width=1\textwidth]{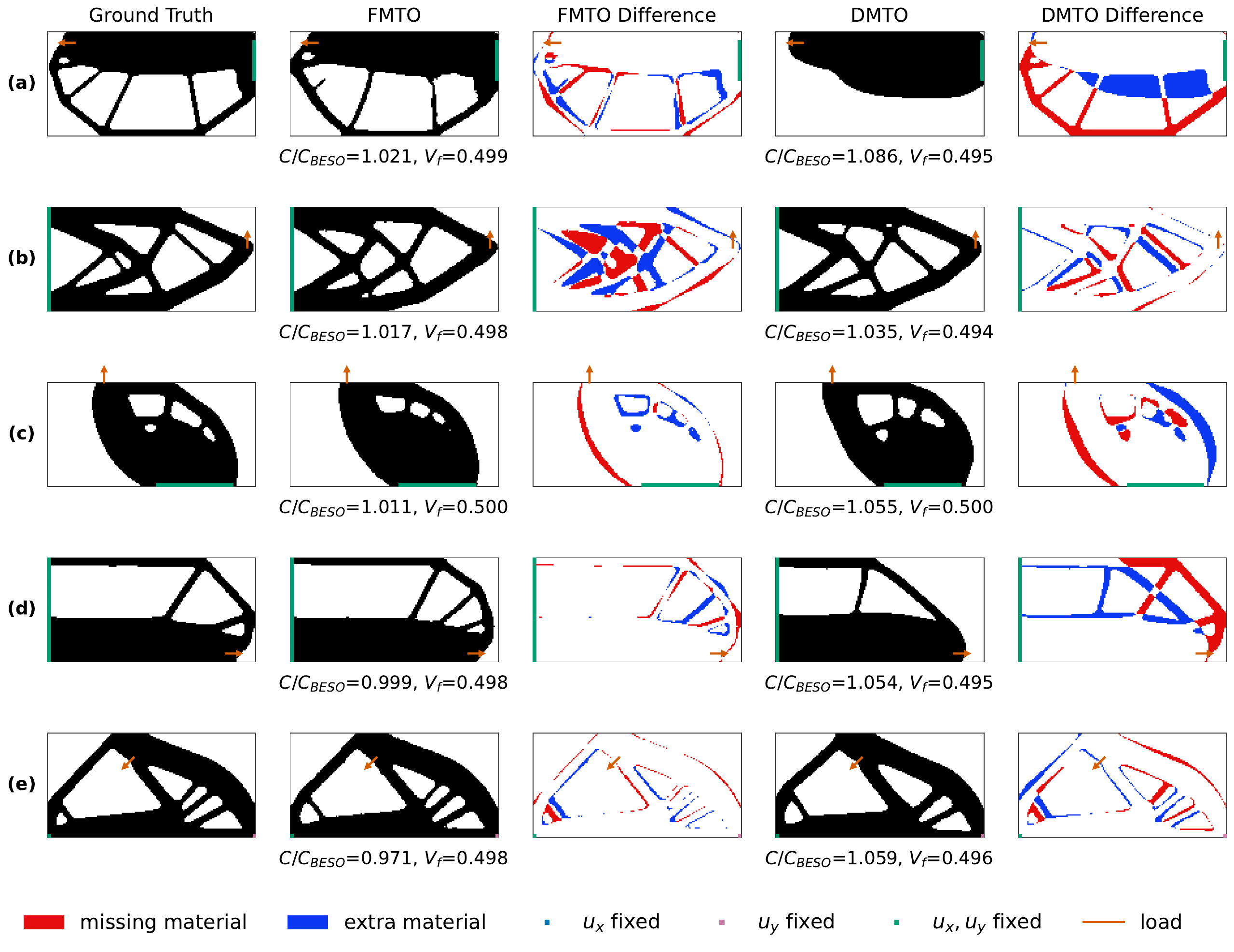}
    \caption{Qualitative comparison of linear FMTO and DMTO under representative validation conditions. Each case shows the BESO reference, the selected generated topology, and the corresponding difference map. Red and blue regions indicate missing and extra material relative to the BESO reference, respectively. For each method, the displayed topology is selected from 100 thresholded candidates satisfying \(V_f \leq 0.5\) using the lowest recomputed \(C/C_{\mathrm{BESO}}\). The reported \(C/C_{\mathrm{BESO}}\) and \(V_f\) correspond to the selected feasible design.}
    \label{fig:fmto_ddpm_qualitative}
\end{figure}

Tables~\ref{tab:fmto_dmto_quality} and~\ref{tab:fmto_dmto_similarity} summarise the quantitative results over 100 validation cases, with 100 generated samples evaluated for each case, resulting in 10000 generated samples per method. All compliance and volume-fraction metrics are computed after thresholding the generated density fields using the same binarisation procedure. As shown in Table~\ref{tab:fmto_dmto_quality}, both FMTO and DMTO achieve mean-best compliance ratios below unity when the best sample is selected from 100 candidates. This indicates that generative sampling can occasionally identify candidates with lower recomputed compliance than the corresponding BESO reference. Accordingly, the best-sample compliance results are interpreted together with volume-fraction feasibility and topology-fidelity metrics, ensuring that the reported improvements correspond to feasible design alternatives rather than numerical artefacts. Therefore, the best-sample statistics are used as an aggregate indicator of potential design improvement, while the case-wise structural validity is further examined through the qualitative comparison in Fig.~\ref{fig:fmto_ddpm_qualitative}.

Despite this constraint, FMTO achieves lower mean-best compliance ratios both with and without enforcing \(V_f\leq0.5\), while also providing a lower median compliance ratio, lower failure rate, higher feasible-sample rate, and smaller volume-fraction deviation. The advantage of FMTO is particularly relevant under the constrained selection criterion, where the best candidate must satisfy \(V_f\leq0.5\), since FMTO maintains a lower mean-best compliance ratio while producing a substantially higher proportion of feasible samples.

Table~\ref{tab:fmto_dmto_similarity} further shows that this improved optimisation behaviour is not obtained by sacrificing topology fidelity. FMTO consistently outperforms DMTO in IoU, Dice, Boundary F1, and Raw MAE, demonstrating better agreement with the BESO reference in both thresholded material layout and continuous density prediction. These results indicate that the FMTO model provides a more reliable generative representation of two-dimensional topology fields, combining improved compliance-related performance, volume-fraction satisfaction, and topology fidelity.

\begin{table}[htbp]
    \centering
    \caption{Optimisation quality and feasibility over 100 validation cases with 100 generated samples per case.}
    \label{tab:fmto_dmto_quality}
    \footnotesize
    \setlength{\tabcolsep}{3.5pt}
    \renewcommand{\arraystretch}{1.15}
    \begin{tabular}{lcccccc}
        \toprule
        Method
        & \makecell{Median\\ \(C/C_{\mathrm{BESO}}\downarrow\)}
        & \makecell{Mean best\\ \(C/C_{\mathrm{BESO}}\downarrow\)}
        & \makecell{Mean best\\ \(V_f \leq 0.5\downarrow\)}
        & \makecell{Fail rate\\ \(C/C_{\mathrm{BESO}}>1.25\downarrow\)}
        & \makecell{Feasible\\ \(V_f \leq 0.5\uparrow\)}
        & \makecell{Mean\\ \(|V_f - V_f^*|\downarrow\)} \\
        \midrule
        FMTO & \textbf{1.0025} & \textbf{0.9588} & \textbf{0.9609} & \textbf{1.54\%} & \textbf{48.07\%} & \textbf{0.0046} \\
        DMTO & 1.0043 & 0.9616 & 0.9763 & 6.10\% & 33.27\% & 0.0094 \\
        \bottomrule
    \end{tabular}
\end{table}

\begin{table}[htbp]
    \centering
    \caption{Topology fidelity with respect to the BESO reference over 100 validation cases, with 100 generated samples per case. IoU, Dice, and Boundary F1 are computed from thresholded topologies, while Raw MAE is computed from the continuous generated density fields. Boundary F1 uses a one-pixel tolerance.}
    \label{tab:fmto_dmto_similarity}
    \begin{tabular}{lcccc}
        \toprule
        Method
        & IoU \(\uparrow\)
        & Dice \(\uparrow\)
        & Boundary F1 \(\uparrow\)
        & Raw MAE \(\downarrow\) \\
        \midrule
        FMTO
        & \textbf{0.8808}
        & \textbf{0.9344}
        & \textbf{0.6792}
        & \textbf{0.0693} \\
        DMTO
        & 0.8013
        & 0.8858
        & 0.4978
        & 0.1191 \\
        \bottomrule
    \end{tabular}
\end{table}

To further examine these differences at the design level, representative validation cases are visualised in Fig.~\ref{fig:fmto_ddpm_qualitative}. For each case and method, the displayed topology is selected from 100 generated samples using the same protocol: candidates must satisfy \(V_f\leq0.5\). Clearly invalid samples, such as those with disconnected load-bearing paths, are excluded by inspection, and the remaining candidate with the lowest recomputed compliance ratio is reported. Overall, FMTO produces designs that more closely resemble the BESO reference in terms of material layout, structural connectivity, and consistency with the prescribed boundary and loading conditions. In cases (b)--(e), the FMTO results preserve the dominant structural features of the reference designs more clearly, whereas the DMTO results show larger deviations, including less continuous material layouts, excessive local material, and distorted structural features. The difference maps provide a consistent visual indication, with FMTO generally exhibiting more localised missing and extra material regions relative to the BESO reference.

The reported compliance ratios in Fig.~\ref{fig:fmto_ddpm_qualitative} further show that the closer agreement with the BESO reference is not obtained at the expense of structural performance. Under the same volume-fraction and visual-validity criteria, the selected FMTO designs achieve compliance ratios close to the BESO reference in all cases, and below \(C/C_{\mathrm{BESO}}=1\) in cases (d) and (e). This indicates that best-of-100 generative sampling can identify structurally plausible feasible candidates with lower recomputed compliance than the corresponding BESO solution. By contrast, the selected DMTO designs generally show higher compliance ratios and larger topological discrepancies, suggesting that DMTO less frequently produces candidates that are simultaneously feasible, structurally reasonable, and low in compliance. These observations are consistent with the aggregate results in Tables~\ref{tab:fmto_dmto_quality} and~\ref{tab:fmto_dmto_similarity}.

\begin{figure}[htbp]
    \centering
    \includegraphics[width=1\textwidth]{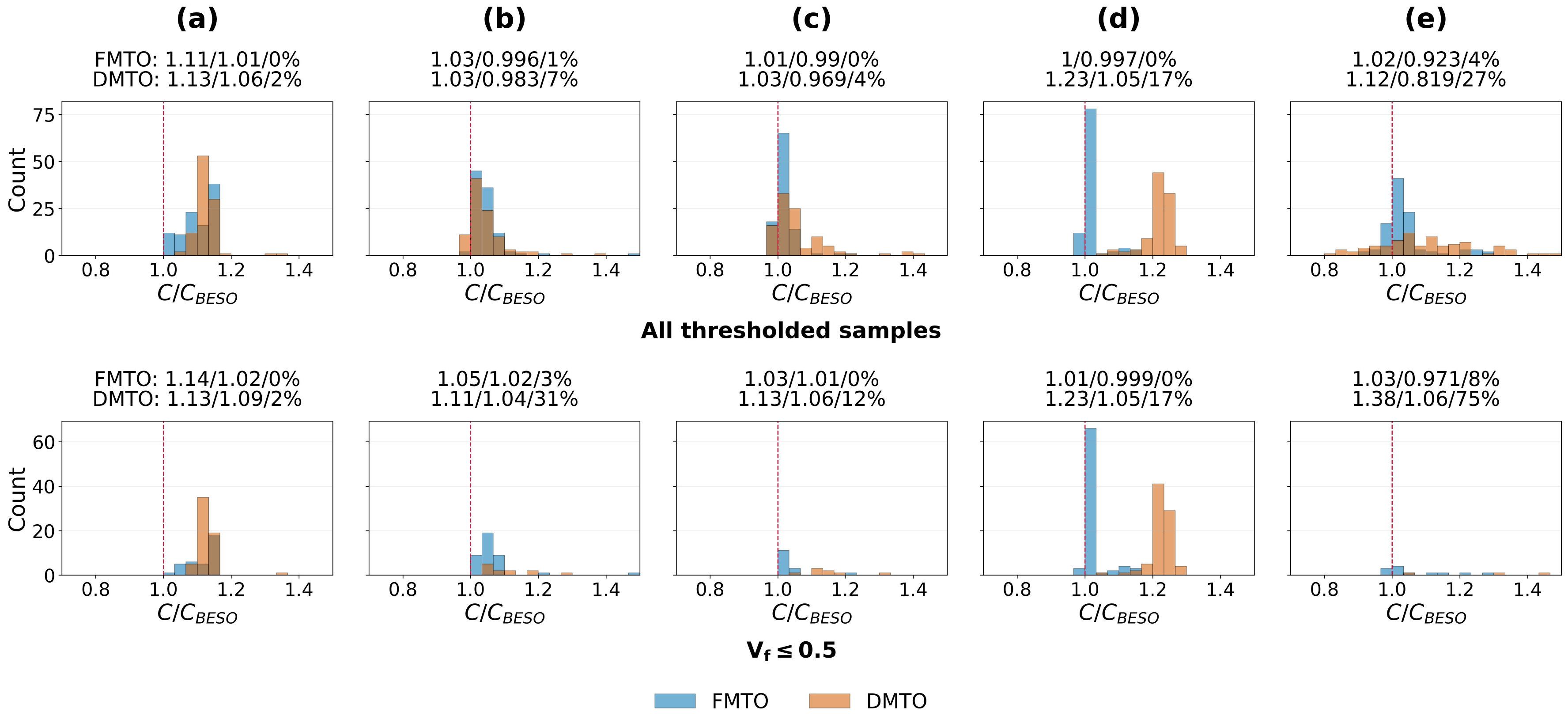}
    \caption{Compliance-ratio distributions of linear FMTO and DMTO over 100 generated samples for five representative validation cases. Each column corresponds to one validation case, labeled (a)--(e). The top row includes all thresholded generated samples, while the bottom row includes only samples satisfying the volume-fraction constraint \(V_f \leq 0.5\). The dashed vertical line marks the BESO reference level \(C/C_{\mathrm{BESO}}=1\). The two values listed above each histogram correspond to linear FMTO and DMTO, respectively, and each entry is reported in the format median compliance ratio / best compliance ratio / failure rate, where the failure rate is the percentage of samples with \(C/C_{\mathrm{BESO}}>1.25\).}
    \label{fig:fmto_dmto_compliance_hist}
\end{figure}

Figure~\ref{fig:fmto_dmto_compliance_hist} further compares the compliance-ratio distributions of the same five validation cases over all 100 generated samples. In the unfiltered distributions, FMTO generally produces more compact distributions around \(C/C_{\mathrm{BESO}}=1\), indicating more stable sampling behaviour. This trend becomes clearer after applying the volume-fraction filter \(V_f\leq0.5\). In the filtered results, FMTO remains more concentrated in all five cases, and, except for case (a), achieves lower median compliance ratios, lower best-sample compliance ratios, and lower failure rates than DMTO. The difference is particularly pronounced in cases (d) and (e), where the DMTO distributions shift towards higher compliance ratios and contain a larger proportion of failed samples. Together, Figs.~\ref{fig:fmto_ddpm_qualitative} and~\ref{fig:fmto_dmto_compliance_hist} show that the advantage of FMTO is reflected both in the selected feasible designs and in the broader distribution of generated candidates satisfying the volume-fraction constraint.

\begin{figure}[htbp]
    \centering
    \includegraphics[width=0.9\textwidth]{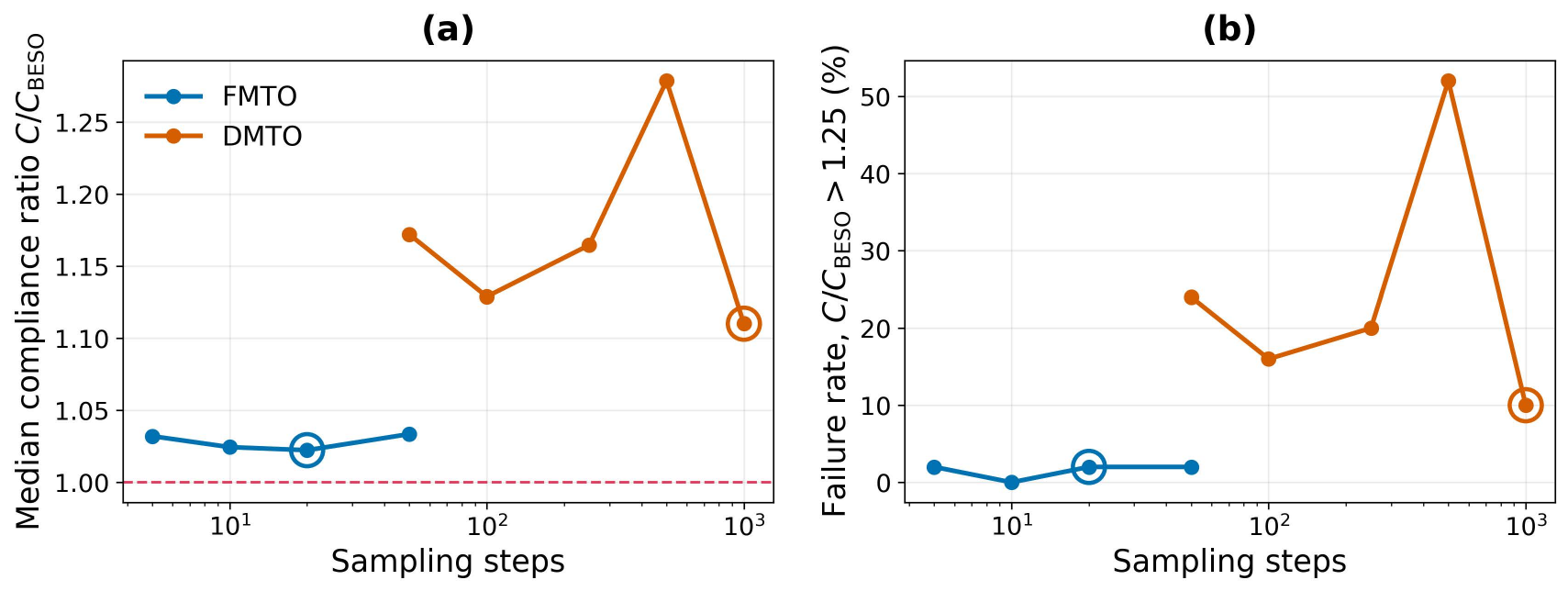}
    \caption{Controlled step-quality analysis for linear FMTO and DMTO on five representative validation cases. For each sampling-step setting, 20 generated samples are evaluated per case. (a) Median compliance ratio \(C/C_{\mathrm{BESO}}\). (b) Failure rate, defined as the percentage of samples with \(C/C_{\mathrm{BESO}}>1.25\). The circled markers indicate the adopted settings: 20 steps for linear FMTO and 1000 steps for DMTO.}
    \label{fig:fmto_dmto_step_quality}
\end{figure}

To justify the adopted sampling-step settings, a controlled step-sensitivity ablation is conducted for both methods, as shown in Fig.~\ref{fig:fmto_dmto_step_quality}. The ablation uses the same five representative validation cases as the qualitative and distributional analyses, so that the sampling-step number is isolated as the main variable. For each step setting, 20 generated samples are evaluated per case. FMTO reaches a stable compliance level with approximately 20 sampling steps, whereas DMTO requires substantially more reverse steps and achieves its best observed performance among the tested settings at 1000 steps. These results support the use of 20 steps for FMTO and 1000 steps for DMTO in the subsequent efficiency comparison.

Based on these settings, Table~\ref{tab:sampling_efficiency} compares the sampling time of the two methods over the same five validation cases, with 100 generated samples per case. The reported timing excludes model loading, visualisation, FEM evaluation, and file input/output (I/O). FMTO requires 0.0177 s per generated sample, whereas DMTO requires 0.8864 s per sample, corresponding to a relative sampling time of 49.9\(\times\). This difference arises from the different sampling mechanisms: FMTO generates samples by integrating a learned deterministic velocity field, which can be accurately approximated using a small number of Euler steps, whereas DMTO follows an iterative reverse-denoising process that typically requires many more steps to obtain stable topology samples.

\begin{table}[htbp]
    \centering
    \caption{Inference sampling efficiency measured over five representative validation cases, with 100 generated samples per case. Timing reports pure sampling only and excludes model loading, visualisation, FEM evaluation, and file I/O.}
    \label{tab:sampling_efficiency}
    \small
    \setlength{\tabcolsep}{5pt}
    \renewcommand{\arraystretch}{1.15}
    \begin{tabular}{lccccc}
        \toprule
        Method & Steps & Cases $\times$ samples & Time/case (s) $\downarrow$ & Time/sample (s) $\downarrow$ & Relative time \(\downarrow\) \\
        \midrule
        FMTO & \textbf{20} & 5 $\times$ 100 & \textbf{1.775} & \textbf{0.0177} & \textbf{1.0\(\times\)} \\
        DMTO & 1000 & 5 $\times$ 100 & 88.638 & 0.8864 & 49.9\(\times\) \\
        \bottomrule
    \end{tabular}
\end{table}

\subsection{Example 2: Trajectory-Aware Path Design under Limited Training Data}\label{Trajectory_Aware_FMTO}

This example evaluates the data efficiency of the proposed trajectory-aware FMTO formulation under a limited training set. While ~\hyperref[Basic_Linear_FMTO]{Example~1} shows that both linear FMTO and DMTO perform well when trained on 5000 BESO-generated TO instances, the dataset size is reduced here to 1000 instances to create a more challenging limited-data setting. This setting is also used to examine whether a moderate trajectory weight gives the best performance, as suggested by the theoretical trade-off between the linear path and the BESO trajectory guide discussed in Section~\ref{subsec:path_velocity_analysis}.

The dataset is divided into training and testing sets with a ratio of 9:1. Each instance contains the converged BESO topology, the global condition vector, the spatial boundary-condition and load fields, and the BESO trajectory used for trajectory-aware path construction. The linear FMTO baseline, DMTO, and trajectory-aware FMTO models with different trajectory weights \(\lambda\) are compared. Here, \(\lambda=0\) corresponds to the linear path, while larger values introduce stronger BESO-trajectory guidance. The material properties, target volume-fraction, FEM setting, and BESO parameters are kept the same as in~\hyperref[Basic_Linear_FMTO]{Example~1}. The trajectory-aware path modifies the training probability path and target velocity field, but does not change the inference ODE. Therefore, it introduces no additional sampling overhead. All FMTO variants are sampled using 20 Euler steps, while DMTO uses 1000 reverse-diffusion steps. Quantitative evaluation is conducted on 50 randomly selected validation cases, with 100 generated topologies per case for each method.

Table~\ref{tab:lambda_quality} summarises the compliance- and feasibility-related results in the limited-data setting. The observed trend is consistent with the theoretical role of \(\lambda\) in Eq.~\eqref{eq:lambda_star}, where the preferred trajectory weight is expected to balance the mismatch of the linear path and the mismatch introduced by the BESO trajectory guide, rather than lie at either extreme. Compared with the linear FMTO baseline \((\lambda=0)\), a moderate trajectory weight improves both constrained best-sample performance and volume feasibility. Among the tested values, \(\lambda=0.25\) achieves the lowest mean-best compliance ratio under \(V_f\leq0.5\), the lowest failure rate, the highest feasible-sample rate, and the smallest mean volume-fraction error. Although larger \(\lambda\) values can reduce the unconstrained mean-best compliance ratio, this improvement is accompanied by substantially weaker volume-constraint satisfaction and higher failure rates. This behaviour agrees with the trade-off described by \(G(\lambda)\), since excessive trajectory guidance increases the contribution of the trajectory-guide mismatch via the \(\lambda^2N_G\) term and restricts the flexibility of the learned generative path.

Table~\ref{tab:lambda_similarity} reports the topology fidelity metrics and shows the same tendency. Trajectory-aware FMTO with \(\lambda=0.25\) obtains the best IoU, Dice, Boundary F1, and Raw MAE, indicating closer agreement with the BESO reference in both thresholded topology layout and continuous density prediction. This indicates that the benefit of a moderate trajectory weight extends beyond compliance performance to topology fidelity. By contrast, DMTO performs substantially worse in both compliance-related and topology-similarity metrics under the same limited-data setting, indicating lower data efficiency and less stable topology generation. Overall, the results identify \(\lambda=0.25\) as the most effective value among those tested and support the theoretically motivated use of moderate trajectory guidance.

\begin{table}[htbp]
    \centering
    \caption{Optimisation quality and feasibility in the limited-data setting with 1000 training samples, evaluated over 50 validation cases with 100 generated samples per case.}
    \label{tab:lambda_quality}
    \footnotesize
    \setlength{\tabcolsep}{3.2pt}
    \renewcommand{\arraystretch}{1.15}
    \begin{tabular}{lcccccc}
        \toprule
        Method
        & \makecell{Median\\ \(C/C_{\mathrm{BESO}}\downarrow\)}
        & \makecell{Mean best\\ \(C/C_{\mathrm{BESO}}\downarrow\)}
        & \makecell{Mean best\\ \(V_f \leq 0.5\downarrow\)}
        & \makecell{Fail rate\\ \(C/C_{\mathrm{BESO}}>1.25\downarrow\)}
        & \makecell{Feasible\\ \(V_f \leq 0.5\uparrow\)}
        & \makecell{Mean\\ \(|V_f - V_f^*|\downarrow\)} \\
        \midrule
        FMTO \(\lambda=0\)    & \textbf{1.0131} & 0.9264 & 0.9481 & 9.80\%  & 62.06\% & 0.0111 \\
        FMTO \(\lambda=0.25\) & 1.0184 & 0.9021 & \textbf{0.9206} & \textbf{8.96\%} & \textbf{70.72\%} & \textbf{0.0104} \\
        FMTO \(\lambda=0.5\)  & 1.0249 & 0.9271 & 0.9572 & 13.94\% & 54.14\% & 0.0106 \\
        FMTO \(\lambda=0.75\) & 1.0286 & 0.8806 & 1.0518 & 19.06\% & 13.48\% & 0.0140 \\
        FMTO \(\lambda=1\)    & 1.0355 & \textbf{0.8660} & 1.1724 & 23.82\% & 1.25\%  & 0.0293 \\
        DMTO                  & \(1.14{\times}10^7\) & 1.1904 & 1.2386 & 82.88\% & 61.71\% & 0.0168 \\
        \bottomrule
    \end{tabular}
\end{table}

\begin{table}[htbp]
    \centering
    \caption{Topology fidelity to the BESO reference in the limited-data setting with 1000 training samples, evaluated over 50 validation cases with 100 generated samples per case. IoU, Dice, and Boundary F1 are computed from thresholded topologies, while Raw MAE is computed from the continuous generated density fields. Boundary F1 uses a one-pixel tolerance.}
    \label{tab:lambda_similarity}
    \footnotesize
    \setlength{\tabcolsep}{5pt}
    \renewcommand{\arraystretch}{1.15}
    \begin{tabular}{lcccc}
        \toprule
        Method
        & IoU \(\uparrow\)
        & Dice \(\uparrow\)
        & Boundary F1 \(\uparrow\)
        & Raw MAE \(\downarrow\) \\
        \midrule
        FMTO \(\lambda=0\)    & 0.7952 & 0.8809 & 0.3394 & 0.1220 \\
        FMTO \(\lambda=0.25\) & \textbf{0.8189} & \textbf{0.8981} & \textbf{0.3660} & \textbf{0.1061} \\
        FMTO \(\lambda=0.5\)  & 0.8060 & 0.8903 & 0.3471 & 0.1223 \\
        FMTO \(\lambda=0.75\) & 0.7098 & 0.8249 & 0.2513 & 0.1866 \\
        FMTO \(\lambda=1\)    & 0.6627 & 0.7896 & 0.2083 & 0.2307 \\
        DMTO                  & 0.4507 & 0.6061 & 0.1305 & 0.3930 \\
        \bottomrule
    \end{tabular}
\end{table}

\begin{figure}[htbp]
    \centering
    \includegraphics[width=1\textwidth]{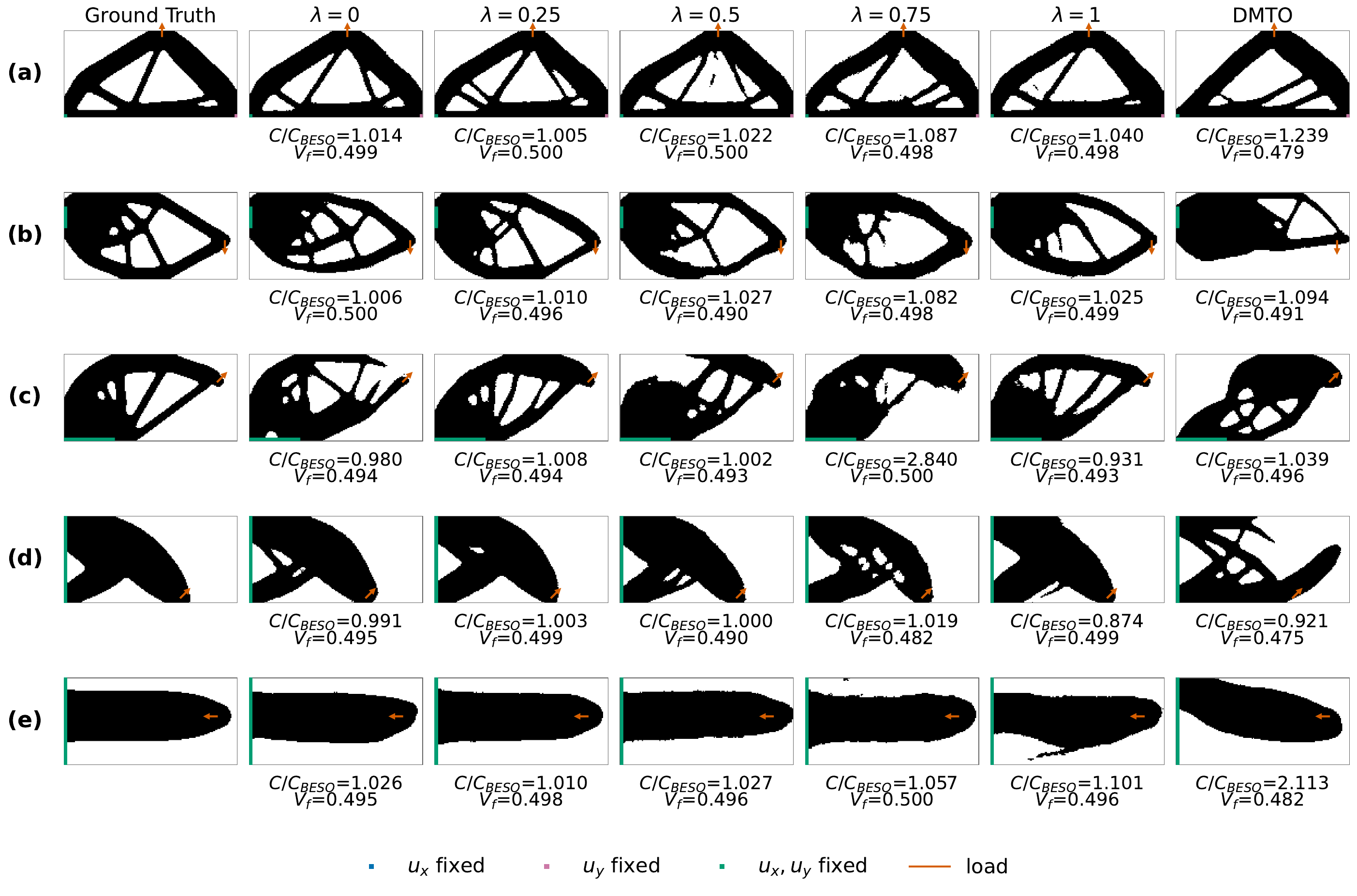}
    \caption{Qualitative comparison of linear FMTO, trajectory-aware FMTO, and DMTO in the limited-data setting with 1000 training samples. Each row shows one validation case, including the BESO reference, FMTO results with \(\lambda\in\{0,0.25,0.5,0.75,1\}\), and the DMTO result. Here, \(\lambda=0\) denotes the linear-path baseline, while larger \(\lambda\) values introduce stronger trajectory guidance. For each method, the displayed topology is selected from 100 thresholded candidates satisfying \(V_f \leq 0.5\), using the candidate whose recomputed compliance ratio is closest to \(C/C_{\mathrm{BESO}}=1\). The reported \(C/C_{\mathrm{BESO}}\) and \(V_f\) correspond to the selected feasible design.}
    \label{fig:fmto_lambda_dmto_1000}
\end{figure}

Figure~\ref{fig:fmto_lambda_dmto_1000} provides a qualitative comparison of the selected feasible designs for five representative validation cases. Consistent with the quantitative results, trajectory-aware FMTO with \(\lambda=0.25\) generally preserves the dominant structural features of the BESO reference more clearly than the linear flow matching baseline and the larger-\(\lambda\) variants. In cases (a)--(c), it produces layouts that remain close to the reference topology while maintaining compliance ratios close to \(C/C_{\mathrm{BESO}}=1\). In cases (d) and (e), where the reference layouts are simpler or more localised, all FMTO variants can generate feasible designs, but stronger trajectory guidance tends to introduce more visible distortions or over-constrained material distributions. These observations also show that compliance ratio alone is insufficient to assess generation quality. For example, in case (c), the linear flow matching result achieves a compliance ratio below unity, but contains visually weak or disconnected structural features. By contrast, \(\lambda=0.25\) provides more stable structural connectivity while maintaining competitive compliance performance, supporting the use of both compliance-based and structural-fidelity-based metrics.

The comparison across \(\lambda\) visually supports the theoretical trade-off
discussed in Section~\ref{subsec:path_velocity_analysis}. A purely linear path ignores the BESO trajectory information, whereas a fully trajectory-dominated path may reduce the flexibility of the learned generative transport. The intermediate setting \(\lambda=0.25\) provides a more effective balance between the linear path and the BESO trajectory guide, improving structural consistency without forcing the generated topology to follow the recorded trajectory too strongly. This trend agrees with Tables~\ref{tab:lambda_quality} and~\ref{tab:lambda_similarity}, where \(\lambda=0.25\) achieves the best overall feasibility and topology fidelity. Compared with the FMTO variants, DMTO shows larger deviations from the BESO reference under the same limited-data setting, particularly in cases (b), (d), and (e), suggesting weaker data efficiency and less stable topology generation when fewer training samples are available.

\subsection{Example 3: Effect of BESO Trajectory Anchor Density}\label{Trajectory_Anchor_Density}

This example investigates the effect of BESO trajectory anchor density on the trajectory-aware FMTO formulation. Based on the results in Example~2, \(\lambda=0.25\) is fixed as the representative trajectory-aware setting, since it gives the best empirical balance between the linear path and the BESO trajectory guide in the limited-data setting. Only the density of the saved BESO trajectory anchors is varied, allowing the effect of trajectory discretisation to be examined independently from the trajectory-guidance strength.

The same limited-data dataset, train--test split, physical parameters, FEM evaluation protocol, and sampling settings as ~\hyperref[Trajectory_Aware_FMTO]{Example~2} are used. The trajectory weight is fixed at \(\lambda=0.25\). Only the trajectory-anchor setting is varied, with \(\alpha=0.05\), \(0.10\), and \(0.25\). Here, \(\alpha\) controls the spacing of saved trajectory anchors, and a smaller \(\alpha\) gives a denser trajectory representation. Quantitative evaluation is conducted on 50 validation cases, with 100 generated topologies per case.

Table~\ref{tab:fmto_anchor_density_quality} summarises the optimisation quality and feasibility results for different BESO trajectory anchor densities. With \(\lambda\) fixed at 0.25, the effect of the anchor density is not monotonic. The setting \(\alpha=0.25\) gives the best median compliance ratio, the best constrained mean-best compliance ratio under \(V_f\leq0.5\), and the lowest failure rate. By contrast, \(\alpha=0.10\) achieves the highest feasible-sample rate and the smallest mean volume-fraction error, indicating better volume-constraint control. The densest setting, \(\alpha=0.05\), achieves the lowest unconstrained mean-best compliance ratio, but also shows the highest failure rate and weaker feasibility. This suggests that overly dense trajectory anchors may improve the chance of finding a low-compliance sample, but can also introduce less stable constraint satisfaction.

Table~\ref{tab:fmto_anchor_density_similarity} reports the topology-similarity metrics. The intermediate setting \(\alpha=0.10\) achieves the best IoU, Dice, Boundary F1, and Raw MAE, showing the closest agreement with the BESO reference in both binary topology layout and continuous density prediction. This indicates that a moderate anchor density provides the most reliable trajectory representation among the tested settings. Sparse anchors may lose important intermediate topology information, whereas overly dense anchors may introduce excessive local trajectory variation. Overall, \(\alpha=0.10\) offers the best balance between trajectory resolution, volume feasibility, and topology fidelity, while \(\alpha=0.25\) remains competitive in compliance-related metrics.

\begin{table}[htbp]
    \centering
    \caption{Optimisation quality and feasibility for different BESO trajectory anchor densities in the limited-data setting with 1000 training samples. All models use $\lambda=0.25$ and are evaluated over 50 validation cases with 100 generated samples per case.}
    \label{tab:fmto_anchor_density_quality}
    \footnotesize
    \setlength{\tabcolsep}{3.2pt}
    \renewcommand{\arraystretch}{1.15}
    \begin{tabular}{lcccccc}
        \toprule
        Method
        & \makecell{Median\\ $C/C_{\mathrm{BESO}}\downarrow$}
        & \makecell{Mean best\\ $C/C_{\mathrm{BESO}}\downarrow$}
        & \makecell{Mean best\\ $V_f \leq 0.5\downarrow$}
        & \makecell{Fail rate\\ \(C/C_{\mathrm{BESO}}>1.25\downarrow\)}
        & \makecell{Feasible\\ $V_f \leq 0.5\uparrow$}
        & \makecell{Mean\\ $|V_f - V_f^*|\downarrow$} \\
        \midrule
        FMTO, anchor density $=0.25$ & \textbf{1.0127} & 0.9205 & \textbf{0.9546} & \textbf{9.10}\% & 34.56\% & 0.0114 \\
        FMTO, anchor density $=0.10$ & 1.0189 & 0.9430 & 0.9667 & 10.88\% & \textbf{44.48}\% & \textbf{0.0092} \\
        FMTO, anchor density $=0.05$ & 1.0368 & \textbf{0.9098} & 0.9626 & 26.98\% & 34.24\% & 0.0124 \\
        \bottomrule
    \end{tabular}
\end{table}

\begin{table}[htbp]
    \centering
    \caption{Topology fidelity to the BESO reference for different BESO trajectory anchor densities in the limited-data setting with 1000 training samples. All models use $\lambda=0.25$ and are evaluated over 50 validation cases with 100 generated samples per case. IoU, Dice, and Boundary F1 are computed from thresholded topologies, while Raw MAE is computed from the continuous generated density fields. Boundary F1 uses a one-pixel tolerance.}
    \label{tab:fmto_anchor_density_similarity}
    \footnotesize
    \setlength{\tabcolsep}{5pt}
    \renewcommand{\arraystretch}{1.15}
    \begin{tabular}{lcccc}
        \toprule
        Method
        & IoU $\uparrow$
        & Dice $\uparrow$
        & Boundary F1 $\uparrow$
        & Raw MAE $\downarrow$ \\
        \midrule
        FMTO, anchor density $=0.25$ & 0.7869 & 0.8778 & 0.3341 & 0.1354 \\
        FMTO, anchor density $=0.10$ & \textbf{0.7980} & \textbf{0.8850} & \textbf{0.3597} & \textbf{0.1266} \\
        FMTO, anchor density $=0.05$ & 0.7630 & 0.8618 & 0.2845 & 0.1455 \\
        \bottomrule
    \end{tabular}
\end{table}

\begin{figure}[htbp]
    \centering
    \includegraphics[width=1\textwidth]{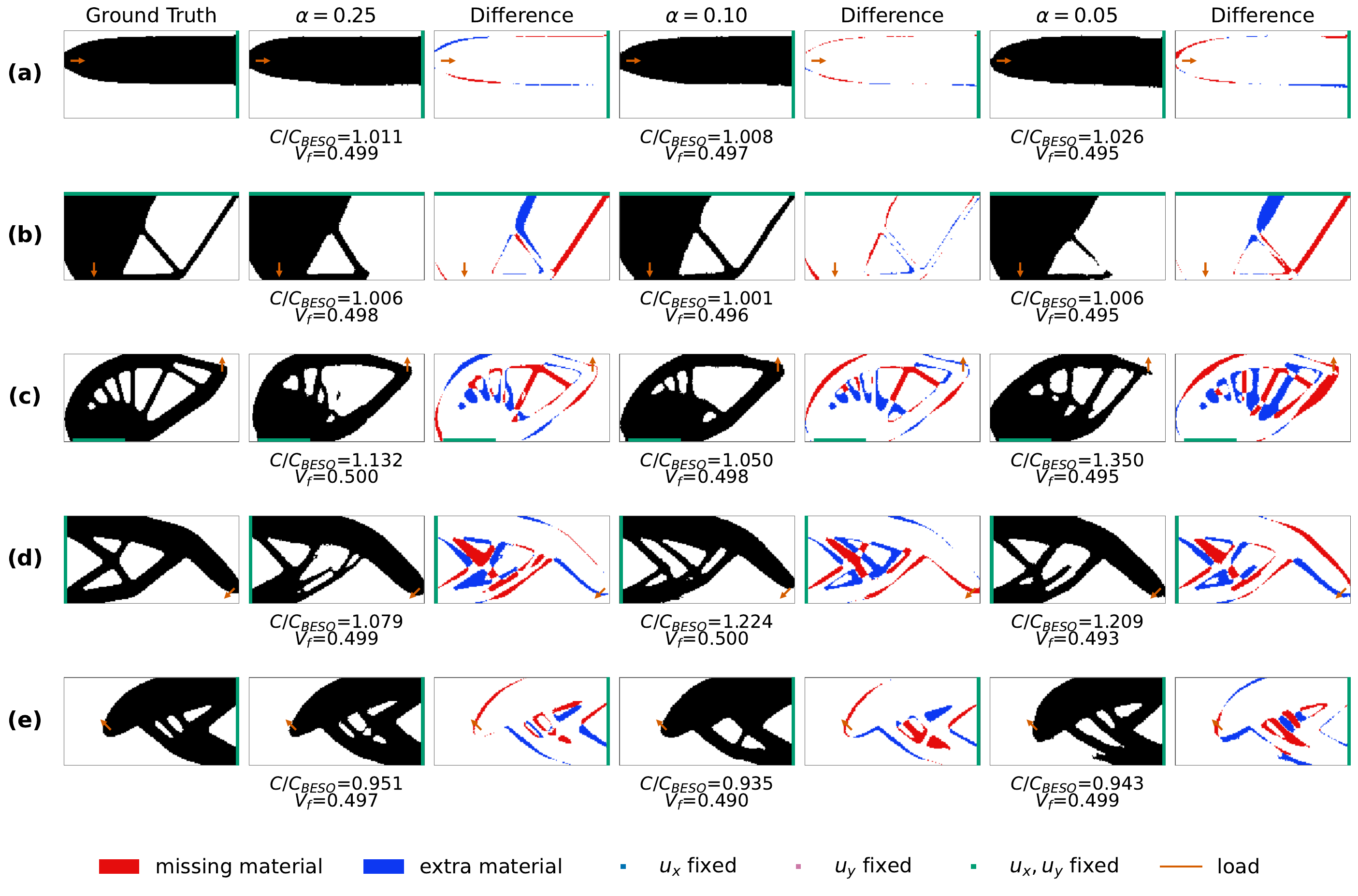}
    \caption{Qualitative comparison of trajectory-aware FMTO with different BESO trajectory anchor densities. Each row shows one validation case, including the BESO reference and generated results for \(\alpha=0.25\), \(0.10\), and \(0.05\). Difference maps use red and blue to indicate missing and extra material relative to the BESO reference. The displayed topology is selected from 100 thresholded candidates, with the reported \(C/C_{\mathrm{BESO}}\) and \(V_f\) corresponding to the selected feasible design.}
    \label{fig:anchor_density_qualitative}
\end{figure}

Figure~\ref{fig:anchor_density_qualitative} presents representative feasible designs generated with different BESO trajectory anchor settings. Since all three models use the same trajectory weight \(\lambda=0.25\) and are trained under the same limited-data setting, their overall topology layouts are visually similar in several cases. The figure therefore mainly highlights local differences induced by the trajectory-anchor discretisation. In cases (a) and (b), all settings recover the main load-transfer paths with only minor local deviations from the BESO reference. In cases (c)--(e), the differences become more visible, particularly around thin members, boundary regions, and local void patterns, where the red and blue regions indicate missing and extra material relative to the reference topology.

Overall, the qualitative comparison suggests that the anchor density affects the local fidelity and stability of the learned trajectory path rather than changing the global design pattern. The intermediate setting \(\alpha=0.10\) generally provides a balanced visual agreement with the BESO reference, preserving the main structural connectivity while avoiding excessive local distortion. By contrast, \(\alpha=0.25\) and \(\alpha=0.05\) can still generate feasible designs, but they show more pronounced local discrepancies in some cases. This indicates that an appropriate trajectory-anchor density is important for representing the BESO optimisation history without either under-resolving the trajectory or introducing excessive local variation into the learned path.

\begin{figure}[htbp]
    \centering
    \includegraphics[width=1\textwidth]{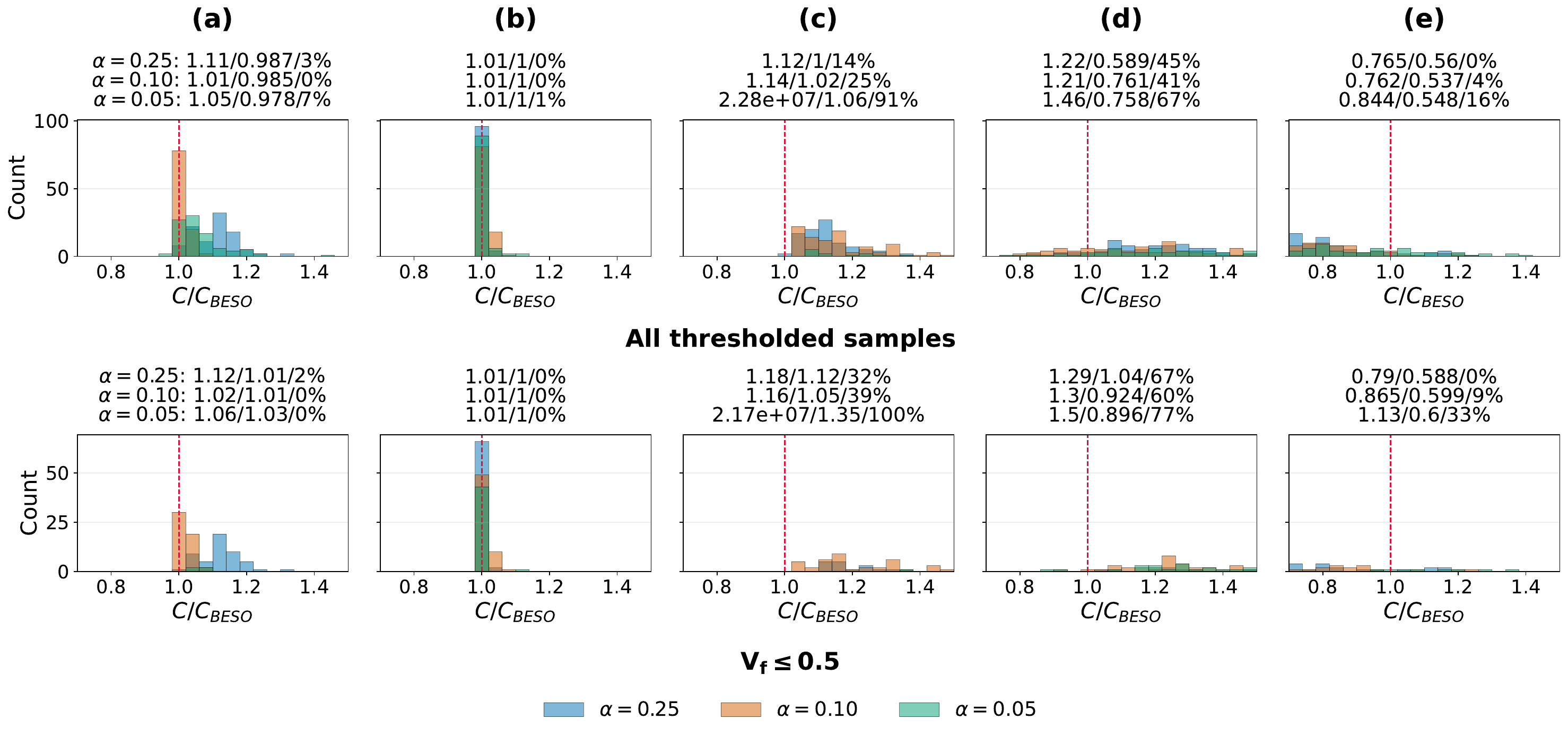}
    \caption{Compliance-ratio distributions of trajectory-aware FMTO with different BESO trajectory anchor densities over 100 generated samples for five representative validation cases. Each column corresponds to one validation case, labeled (a)--(e). The top row includes all thresholded generated samples, while the bottom row includes only samples satisfying the volume-fraction constraint \(V_f \leq 0.5\). The dashed vertical line marks the BESO reference level \(C/C_{\mathrm{BESO}}=1\). The values listed above each histogram are reported in the format median compliance ratio / best compliance ratio / failure rate, where the failure rate is the percentage of samples with \(C/C_{\mathrm{BESO}}>1.25\).}
    \label{fig:anchor_density_compliance_hist}
\end{figure}

To further examine the stochastic stability of the generated samples, Fig.~\ref{fig:anchor_density_compliance_hist} compares the compliance-ratio distributions for the same five representative validation cases. Cases (a) and (b) show compact distributions close to the BESO reference for all three anchor densities, indicating that the anchor density has limited influence when the generation problem is relatively stable. In contrast, cases (c)--(e) exhibit broader distributions and heavier high-compliance tails, especially for \(\alpha=0.05\). This suggests that overly dense trajectory anchoring may introduce excessive local path variation for more challenging design conditions, even when visually feasible samples can still be generated.

The histograms also show that the best compliance ratio alone does not fully characterise sampling robustness. Although \(\alpha=0.05\) can achieve low best-sample compliance in some cases, it also produces broader distributions and higher failure rates. Therefore, Fig.~\ref{fig:anchor_density_compliance_hist} complements the qualitative comparison by revealing the sample-level stability of each anchor-density setting. Together with the aggregate results in Tables~\ref{tab:fmto_anchor_density_quality} and~\ref{tab:fmto_anchor_density_similarity}, these observations support the use of the intermediate anchor setting \(\alpha=0.10\), which provides strong topology fidelity while avoiding the instability associated with overly dense trajectory discretisation.

\FloatBarrier

\subsection{Example 4: Three-Dimensional Applicability, Volume Feasibility, and Topology Fidelity}
\label{ThreeD_FMTO}

This example examines the applicability of the proposed FMTO framework to
three-dimensional TO. A dataset of 2000 BESO-generated 3D TO instances is used,
with 1800 instances for training and 200 instances for testing. Each instance
contains the converged BESO reference topology, the global condition vector, and
the three-dimensional boundary-condition and load fields. The voxel resolution is
\(48\times24\times24\). Linear FMTO, trajectory-aware FMTO with
\(\lambda=0.25\), and DMTO are compared. The trajectory weight is adopted from
\hyperref[Trajectory_Aware_FMTO]{Example~2}.

Since full 3D FEM compliance evaluation is computationally expensive, the
quantitative comparison focuses on voxel-level topology fidelity and volume-fraction feasibility. The target volume fraction is reduced to \(V_f^*=0.3\), and the volume-fraction failure rate of \(V_f/V_f^*>1.05\) is adopted. Table~\ref{tab:fmto_dmto_3d_fidelity_feasibility} summarises the results over 100 validation cases, with 50 generated samples per case. Trajectory-aware FMTO achieves the best performance across all reported metrics, with the highest IoU, Dice, Boundary F1, and feasible-sample rate, as well as the lowest Raw MAE, volume-fraction failure rate, and mean volume-fraction error. Linear FMTO also consistently outperforms DMTO. These results show that BESO-trajectory guidance improves both voxel-level topology fidelity and volume-fraction control in the three-dimensional setting.

\begin{table}[htbp]
    \centering
    \caption{Topology fidelity and volume-fraction feasibility of 3D generated
    topologies over 100 validation cases, with 50 generated samples per case.
    IoU, Dice, Boundary F1, and feasibility metrics are computed from thresholded
    voxel fields, while Raw MAE is computed from the continuous generated density
    fields. Boundary F1 uses a one-voxel tolerance.}
    \label{tab:fmto_dmto_3d_fidelity_feasibility}
    \footnotesize
    \setlength{\tabcolsep}{3.5pt}
    \renewcommand{\arraystretch}{1.15}
    \begin{tabular}{lccccccc}
        \toprule
        Method
        & IoU \(\uparrow\)
        & Dice \(\uparrow\)
        & \makecell{Boundary\\ F1 \(\uparrow\)}
        & \makecell{Raw\\ MAE \(\downarrow\)}
        & \makecell{Volume fail rate\\ \(V_f/V_f^*>1.05\downarrow\)}
        & \makecell{Feasible\\ \(V_f\leq0.3\uparrow\)}
        & \makecell{Mean\\ \(|V_f-V_f^*|\downarrow\)} \\
        \midrule
        Linear FMTO
        & 0.7769 & 0.8709 & 0.8330 & 0.0829
        & 4.16\% & 47.98\% & 0.0068 \\
        Trajectory-aware FMTO
        & \textbf{0.8698} & \textbf{0.9284} & \textbf{0.9514}
        & \textbf{0.0474} & \textbf{0.74\%} & \textbf{55.30\%}
        & \textbf{0.0043} \\
        DMTO
        & 0.6963 & 0.8136 & 0.7304 & 0.1154
        & 5.96\% & 42.02\% & 0.0072 \\
        \bottomrule
    \end{tabular}
\end{table}

\begin{figure}[htbp]
    \centering
    \includegraphics[width=0.7\textwidth]{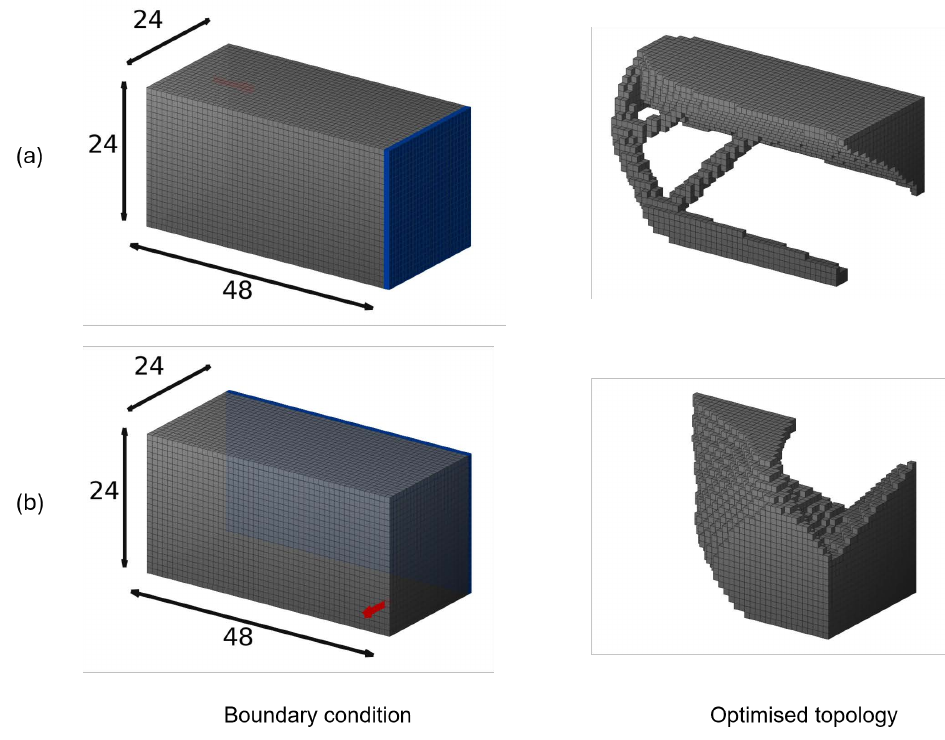}
    \caption{Prescribed 3D design domains and corresponding BESO reference
    topologies for two representative validation cases. The left column shows the
    initial design domains with voxel resolution \(48\times24\times24\), where
    blue regions indicate fixed supports and red arrows denote prescribed point
    loads. The right column shows the corresponding converged BESO reference
    topologies.}
    \label{fig:initial_domain_3d_cases}
\end{figure}

\begin{figure}[htbp]
    \centering
    \includegraphics[width=\textwidth]{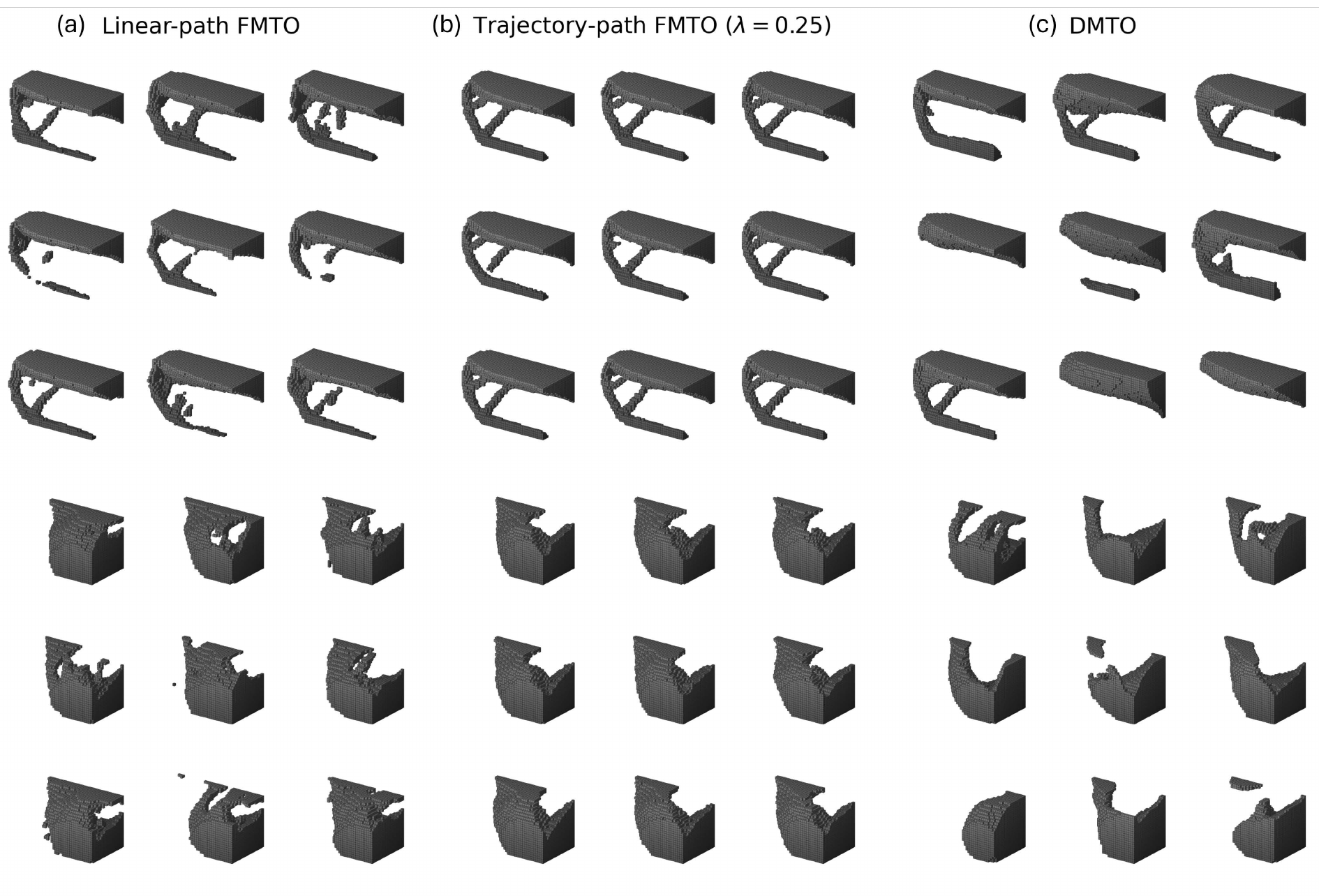}
    \caption{Qualitative comparison of 3D topology samples generated by
    (a) linear FMTO, (b) trajectory-aware FMTO with \(\lambda=0.25\), and
    (c) DMTO for the two validation cases defined in
    Fig.~\ref{fig:initial_domain_3d_cases}. For each method, the upper and lower
    \(3\times3\) blocks correspond to the first and second validation cases,
    respectively.}
    \label{fig:three_model_3d_val015_samples}
\end{figure}

Figure~\ref{fig:initial_domain_3d_cases} shows the prescribed design domains and
BESO reference topologies for two representative validation cases with different
support regions and load locations.
Figure~\ref{fig:three_model_3d_val015_samples} compares nine generated samples
from each method under the two conditions. Trajectory-aware FMTO produces more
consistent material layouts and more clearly preserves the dominant features of
the BESO references. Linear FMTO captures the overall topology patterns but
occasionally produces fragmented local features, while DMTO exhibits larger
sample-to-sample variation and greater deviations from the reference topologies.
These observations are consistent with
Table~\ref{tab:fmto_dmto_3d_fidelity_feasibility}.

For the 100 validation cases with 50 samples per case, linear FMTO and
trajectory-aware FMTO required approximately 148~s and 150~s, respectively,
whereas DMTO required approximately 6181~s. Thus, DMTO was approximately
\(41\times\) slower than the FMTO variants, while trajectory-aware path
construction introduced negligible inference-time overhead.

\FloatBarrier

\section{Discussion and Conclusions}\label{sec5}

This work proposes a trajectory-aware flow matching-based topology optimisation (FMTO) framework for conditional topology generation under prescribed physical and design conditions. A basic linear FMTO formulation is first developed by constructing a probability path between a Gaussian source field and the BESO reference topology and training a conditional velocity field for ODE-based topology generation. Although this endpoint-based path provides a simple and efficient baseline, its intermediate states contain no information about the mechanics-driven material evolution that produces the target topology. Probability-path construction is therefore treated as a central design problem in FMTO. To address this limitation, a BESO-guided trajectory path is introduced. Instead of using only the converged topology as the endpoint target, the trajectory-aware formulation uses volume-fraction-indexed intermediate BESO states to construct the probability path and target velocity field. The resulting framework generates multiple candidate topologies from different source samples while requiring only the prescribed physical and design conditions during inference.

A key theoretical contribution of this work is the path--velocity mismatch analysis developed to evaluate trajectory-aware path construction. By accounting for both centreline and velocity mismatch, the analysis characterises the trade-off between endpoint transport and BESO trajectory guidance and explains why a moderate trajectory weight may outperform both the endpoint-only path and strongly trajectory-dominated paths. Under the stated assumptions, the mismatch is further connected to velocity-field approximation error, terminal generation error, and the probability of generating high-compliance failure samples. This analysis provides a mechanics-informed explanation of how probability-path design affects the stability and quality of generative TO.

The numerical examples demonstrate the effectiveness of the proposed framework. In the basic two-dimensional setting, linear FMTO achieves better compliance-related performance, volume-fraction satisfaction, and topology fidelity than the diffusion-model baseline, while requiring substantially fewer sampling steps. Under the same recomputed finite-element evaluation protocol, best-of-samples generation also identifies candidate structures with compliance comparable to, and in some cases lower than, the corresponding reference topology. These results show that FMTO can efficiently generate diverse and competitive design alternatives under prescribed physical conditions.

In the limited-data setting, trajectory-aware FMTO improves the robustness and data efficiency of topology generation. Among the tested trajectory weights, a moderate value of \(\lambda=0.25\) gives the best overall balance, achieving improved constrained compliance performance, a lower failure rate, better volume feasibility, and higher topology fidelity to the BESO reference. This trend is consistent with the path--velocity mismatch analysis, supporting the use of moderate BESO trajectory guidance. The anchor-density study further shows that trajectory representation affects local fidelity and sampling stability, with an intermediate anchor setting providing the most reliable balance between trajectory resolution, volume feasibility, and topology fidelity. The three-dimensional examples further demonstrate the applicability of the framework to voxel-based TO, where trajectory-aware FMTO qualitatively produces more consistent material layouts and better preserves the dominant load-transfer features than the linear-path and diffusion-based baselines.

Overall, the proposed framework demonstrates that generative TO can benefit not only from efficient ODE-based sampling, but also from mechanics-informed probability-path construction. The key contribution of trajectory-aware FMTO is the conversion of conventional TO optimisation histories into path and velocity supervision for flow matching, providing a direct link between mechanics-driven material evolution and generative topology synthesis.

Nevertheless, the current framework still has limitations. The present study focuses mainly on linear elastic compliance-minimisation problems, and the generated continuous density fields are evaluated after fixed-threshold binarisation. The trajectory-aware path is constructed from BESO histories and volume-fraction-indexed intermediate states, while other types of optimisation trajectories, intermediate physical fields, and path parameterisations remain unexplored. The trajectory weight and anchor density are selected empirically in the numerical studies, and the theoretical analysis relies on a hypothetical ideal generative path and standard error-propagation assumptions. In addition, the three-dimensional validation is still limited to moderate-resolution voxel-based cases and should be extended to larger-scale problems with full 3D compliance evaluation and connectivity-aware validity metrics.

Future work will investigate more general forms of trajectory-aware flow matching for TO and broader generative design problems. Although this study uses BESO trajectories, the same principle can be extended to other conventional TO formulations, including SIMP density-evolution histories, level-set boundary evolution, MMC parameter trajectories, and BESO sensitivity histories. The intermediate information used to guide the flow path also need not be restricted to volume-fraction-indexed topology snapshots. Possible alternatives include compliance-sensitivity fields, adjoint-gradient information, pseudo-sensitivity fields, displacement fields, strain-energy histories, stress fields, constraint-violation maps, and multi-physics response fields. This direction is consistent with recent sensitivity-conditioned Bernoulli flow matching for TO, where sensitivity and pseudo-sensitivity fields are used as informative conditioning signals for improved generalisation~\cite{rashed2026generalization}. The trajectory-aware flow matching framework could also be combined with advanced generative modelling approaches, including optimal-transport flow matching~\cite{lipman2022flow}, rectified flow~\cite{liu2022flow}, stochastic interpolants~\cite{albergo2025stochastic}, neural operators, transformer-based condition encoders, and diffusion--flow hybrid models. These developments may further extend the proposed framework to more complex TO problems involving multiple objectives, uncertainty, nonlinear responses, and coupled multi-physics constraints.

\section*{CRediT authorship contribution statement}

\textbf{Shusheng Xiao}: Conceptualisation, Methodology, Coding, Formal analysis, Writing -- original draft.
\textbf{Jinshuai Bai}: Conceptualisation, Methodology, Formal analysis, Supervision, Funding acquisition, Writing -- review \& editing.
\textbf{Hyogu Jeong}: Formal analysis, Writing --review \& editing.
\textbf{Yunfei Xi}: Formal analysis, Funding acquisition, Writing --review \& editing.
\textbf{Yilin Gui}: Formal analysis, Writing --review \& editing.
\textbf{YuanTong Gu}: Conceptualisation, Formal analysis, Supervision, Project administration, Funding acquisition, Writing --review \& editing.

\section*{Declaration of competing interest}

The authors declare that they have no known competing financial interests or personal relationships that could have appeared to influence the work reported in this paper.

\section*{Acknowledgements}

Support from the National Natural Science Foundation of China (Grant no. 12502234) and China Postdoctoral Science Foundation (Grant No. 2025M781785) are acknowledged (J. Bai). Support from the Australian Research Council Mid-Career Industry Fellowship (IM230100132) is gratefully acknowledged (Y.T. Gu). Support from a Queensland University of Technology Australian Postgraduate Research Award (QUTPRA).

\section*{Data availability}

Data will be made available on request.

\appendix

\section{Derivation of the path--velocity mismatch}
\label{app:path_velocity_mismatch}

This appendix provides the derivation of the path--velocity mismatch used in
Section~\ref{subsec:path_velocity_analysis}. Let \(\boldsymbol{m}^{*}_i(t)\) denote a hypothetical
ideal generative centreline. For the endpoint-based linear path, the centreline is
\(\boldsymbol{m}^{0}_i(t)=\boldsymbol{\rho}^{\mathrm{ref}}_i\). Its position and velocity
mismatches relative to the ideal centreline are

\begin{equation}
\boldsymbol{b}_i(t)
=
\boldsymbol{\rho}^{\mathrm{ref}}_i
-
\boldsymbol{m}^{*}_i(t),
\qquad
\boldsymbol{a}_i(t)
=
-
\dot{\boldsymbol{m}}^{*}_i(t).
\label{eq:app_linear_mismatch}
\end{equation}

For the BESO trajectory guide, the corresponding mismatches are

\begin{equation}
\boldsymbol{e}_i(t)
=
\boldsymbol{q}_i(t)
-
\boldsymbol{m}^{*}_i(t),
\qquad
\boldsymbol{r}_i(t)
=
\dot{\boldsymbol{q}}_i(t)
-
\dot{\boldsymbol{m}}^{*}_i(t).
\label{eq:app_trajectory_mismatch}
\end{equation}

Using Eq.~\eqref{eq:guided_center}, the guided centreline satisfies

\begin{equation}
\boldsymbol{m}^{\lambda}_i(t)
-
\boldsymbol{m}^{*}_i(t)
=
(1-\lambda)\boldsymbol{b}_i(t)
+
\lambda\boldsymbol{e}_i(t),
\label{eq:app_guided_position_mismatch}
\end{equation}

\noindent and

\begin{equation}
\dot{\boldsymbol{m}}^{\lambda}_i(t)
-
\dot{\boldsymbol{m}}^{*}_i(t)
=
(1-\lambda)\boldsymbol{a}_i(t)
+
\lambda\boldsymbol{r}_i(t).
\label{eq:app_guided_velocity_mismatch}
\end{equation}

Because flow matching learns a velocity field, the path quality is measured using both centreline
position and induced velocity. The combined mismatch is

\begin{equation}
G(\lambda)
=
\mathbb{E}_{i,t}
\left[
\left\|
\boldsymbol{m}^{\lambda}_i(t)
-
\boldsymbol{m}^{*}_i(t)
\right\|_2^2
+
\eta
\left\|
\dot{\boldsymbol{m}}^{\lambda}_i(t)
-
\dot{\boldsymbol{m}}^{*}_i(t)
\right\|_2^2
\right],
\qquad
\eta>0.
\label{eq:app_mismatch_definition}
\end{equation}

Substituting Eqs.~\eqref{eq:app_guided_position_mismatch} and
\eqref{eq:app_guided_velocity_mismatch} into Eq.~\eqref{eq:app_mismatch_definition} gives

\begin{equation}
G(\lambda)
=
(1-\lambda)^2B_G
+
\lambda^2N_G
+
2\lambda(1-\lambda)C_G,
\label{eq:app_mismatch_expanded}
\end{equation}

\noindent where

\begin{equation}
B_G
=
\mathbb{E}_{i,t}
\left[
\|\boldsymbol{b}_i(t)\|_2^2
+
\eta\|\boldsymbol{a}_i(t)\|_2^2
\right],
\label{eq:app_BG}
\end{equation}

\begin{equation}
N_G
=
\mathbb{E}_{i,t}
\left[
\|\boldsymbol{e}_i(t)\|_2^2
+
\eta\|\boldsymbol{r}_i(t)\|_2^2
\right],
\label{eq:app_NG}
\end{equation}

\noindent and

\begin{equation}
C_G
=
\mathbb{E}_{i,t}
\left[
\left\langle
\boldsymbol{b}_i(t),
\boldsymbol{e}_i(t)
\right\rangle
+
\eta
\left\langle
\boldsymbol{a}_i(t),
\boldsymbol{r}_i(t)
\right\rangle
\right].
\label{eq:app_CG}
\end{equation}

\noindent Since \(G(0)=B_G\), the trajectory-aware path improves the mismatch relative to the linear
path when \(G(\lambda)<G(0)\). For \(\lambda>0\), this condition becomes

\begin{equation}
\lambda
\left(
B_G+N_G-2C_G
\right)
<
2
\left(
B_G-C_G
\right).
\label{eq:app_improvement_condition}
\end{equation}

\noindent Thus, if the BESO guide contains useful path information and is not over-weighted, there
exists a positive range of \(\lambda\) for which the trajectory-aware path has a smaller
path--velocity mismatch than the linear path. In the non-degenerate case, the unconstrained
minimiser is

\begin{equation}
\lambda^{*}
=
\frac{
B_G-C_G
}{
B_G+N_G-2C_G
},
\label{eq:app_lambda_star}
\end{equation}

\noindent and the constrained value is obtained by projecting this expression onto \([0,1]\).

\section{Error propagation and failure probability}
\label{app:error_propagation}

This appendix provides the error-propagation argument connecting path--velocity mismatch with
terminal generation stability. The population velocity-field approximation error for the
trajectory-aware path is

\begin{equation}
\epsilon_{\lambda}
=
\int_{0}^{1}
\mathbb{E}_{\boldsymbol{x}_t\sim p_t^{\lambda}}
\left[
\left\|
\boldsymbol{v}^{\lambda}_{\theta}
\left(
\boldsymbol{x}_t,
t,
\boldsymbol{c}
\right)
-
\boldsymbol{u}^{\lambda}_{t}
\left(
\boldsymbol{x}_t,
\boldsymbol{c}
\right)
\right\|_2^2
\right]
\mathrm{d}t,
\label{eq:app_velocity_error}
\end{equation}

\noindent where \(p_t^{\lambda}\) is the distribution induced by the trajectory-aware path at time
\(t\). To relate this learning error to path construction, we use the learning-error transfer
assumption

\begin{equation}
\epsilon_{\lambda}
\leq
A G(\lambda)
+
R_{\lambda},
\label{eq:app_learning_transfer}
\end{equation}

\noindent where \(A>0\) is a transfer constant and \(R_{\lambda}\) collects finite-data,
finite-capacity, optimisation, and training-randomness errors. Under controlled comparisons using
the same architecture, dataset size, optimiser, and training schedule, the residual terms are
assumed to be comparable,

\begin{equation}
R_{\lambda}
\approx
R_{0}.
\label{eq:app_residual_comparable}
\end{equation}

Let \(\hat{\boldsymbol{\rho}}^{\lambda}_{\theta}(\boldsymbol{c},\boldsymbol{z})\) denote the
generated topology for condition \(\boldsymbol{c}\) and source noise
\(\boldsymbol{z}\sim\mathcal{N}(\boldsymbol{0},\boldsymbol{I})\). The terminal generation error is

\begin{equation}
D_{\lambda}(\boldsymbol{c})
=
\mathbb{E}_{\boldsymbol{z}}
\left[
\left\|
\hat{\boldsymbol{\rho}}^{\lambda}_{\theta}
\left(
\boldsymbol{c},
\boldsymbol{z}
\right)
-
\boldsymbol{\rho}^{\mathrm{ref}}(\boldsymbol{c})
\right\|_2^2
\right].
\label{eq:app_terminal_error}
\end{equation}

If the learned velocity field satisfies a Lipschitz-type condition, a Gronwall-type ODE stability
bound gives

\begin{equation}
D_{\lambda}(\boldsymbol{c})
\leq
K\epsilon_{\lambda}
+
K_{\mathrm{num}}h^{2p},
\label{eq:app_ode_stability}
\end{equation}

\noindent where \(K\) is an error-propagation constant, \(h\) is the integration step size, \(p\) is
the order of the numerical solver, and \(K_{\mathrm{num}}h^{2p}\) is the numerical integration
error. When the same solver and step size are used, this term is comparable across different path
constructions.

To connect the terminal error with engineering stability, define the bad-structure set under
condition \(\boldsymbol{c}\) as

\begin{equation}
\mathcal{B}_{\boldsymbol{c}}
=
\left\{
\boldsymbol{\rho}:
\frac{
C(\boldsymbol{\rho},\boldsymbol{c})
}{
C_{\mathrm{BESO}}(\boldsymbol{c})
}
>
\tau
\right\},
\label{eq:app_bad_structure_set}
\end{equation}

\noindent where \(\tau>1\) is a prescribed failure threshold. Assume that every topology in this set
is separated from the reference topology by at least \(\delta\),

\begin{equation}
\boldsymbol{\rho}\in\mathcal{B}_{\boldsymbol{c}}
\Rightarrow
\left\|
\boldsymbol{\rho}
-
\boldsymbol{\rho}^{\mathrm{ref}}(\boldsymbol{c})
\right\|_2
\geq
\delta.
\label{eq:bad_structure_separation}
\end{equation}

Applying Markov's inequality to the squared terminal error gives

\begin{equation}
P^{\lambda}_{\mathrm{fail}}
\left(
\mathcal{B}_{\boldsymbol{c}}
\right)
\leq
\frac{
D_{\lambda}(\boldsymbol{c})
}{
\delta^2
}.
\label{eq:app_failure_bound}
\end{equation}

\noindent Therefore, under the stated assumptions and comparable residual errors, a smaller
path--velocity mismatch can tighten the upper bounds on the velocity-field approximation error,
terminal generation error, and failure probability. This argument is interpretive and does not
constitute an unconditional guarantee.

\bibliographystyle{elsarticle-num-names}
\bibliography{main.bib}
\end{document}